\documentclass[conference]{IEEEtran}
\IEEEoverridecommandlockouts
\usepackage{graphicx}
\usepackage[hyphens]{url}
\usepackage{algorithmic}
\usepackage{textcomp}
\usepackage{comment}
 \usepackage{multirow}
\usepackage[ruled,vlined]{algorithm2e}
\SetKwInOut{Parameter}{Parameter}
\SetKwInOut{Input}{Input}
\SetKwInOut{Output}{Output}
\usepackage{subcaption}
\usepackage{tabularx,booktabs}
\usepackage{amsmath}
\usepackage[table,xcdraw]{xcolor}
\usepackage{pgfplots}
\usepackage{makecell}
\usepackage{cite}
\usepackage{amsmath,amssymb,amsfonts}
\usepackage{textcomp}
\begin{document}

\title{Adversarial Examples in Deep Learning for Multivariate Time Series Regression}


\author{\IEEEauthorblockN{Gautam Raj Mode and Khaza Anuarul Hoque}
\IEEEauthorblockA{\textit{Department of Electrical Engineering \& Computer Science} \\
\textit{University of Missouri, Columbia, MO, USA}\\
gmwyc@mail.missouri.edu, hoquek@missouri.edu}

}

\maketitle

\begin{abstract}
Multivariate time series (MTS) regression tasks are common in many real-world data mining applications including finance, cybersecurity, energy, healthcare, prognostics, and many others. Due to the tremendous success of deep learning (DL) algorithms in various domains including image recognition and computer vision, researchers started adopting these techniques for solving MTS data mining problems, many of which are targeted for safety-critical and cost-critical applications. Unfortunately, DL algorithms are known for their susceptibility to adversarial examples which also makes the DL regression models for MTS forecasting also vulnerable to those attacks. To the best of our knowledge, no previous work has explored the vulnerability of DL MTS regression models to adversarial time series examples, which is an important step, specifically when the forecasting from such models is used in safety-critical and cost-critical applications. In this work, we leverage existing adversarial attack generation techniques from the image classification domain and craft adversarial multivariate time series examples for three state-of-the-art deep learning regression models, specifically Convolutional Neural Network (CNN), Long Short-Term Memory (LSTM), and Gated Recurrent Unit (GRU). We evaluate our study using Google stock and household power consumption dataset. The obtained results show that all the evaluated DL regression models are vulnerable to adversarial attacks, transferable, and thus can lead to catastrophic consequences in safety-critical and cost-critical domains, such as energy and finance. 

\end{abstract}

\section{Introduction}
\label{sec:introduction}
Time series forecasting is an important problem in data mining with many real-world applications including finance \cite{sezer2020financial,gan2020machine,sirignano2019universal,lee2019multimodal}, weather forecasting \cite{salman2015weather,hossain2015forecasting}, power consumption monitoring \cite{khan2018load,chan2019deep}, industrial maintenance \cite{fontes2016pattern,lei2019fault}, occupancy monitoring in smart buildings \cite{zou2018towards,zhang2018thermal}, and many others. Recently, deep learning (DL) models showed tremendous success in analyzing time series data~\cite{sezer2020financial,fawaz2019deep} when compared to the other traditional methods. This is due to the fact that DL models can automatically learn complex mappings from multiple inputs to outputs. Interestingly, DL models can be easily fooled by adversarial examples~\cite{goodfellow2014explaining,kurakin2016adversarial}. From the perspective of image processing or computer vision, an adversarial example can be an image formed by making small perturbations (insignificant to the human eye) to an example image. Another interesting fact is that the adversarial examples can often transfer from one model to another model, known as \emph{black-box attacks}, which means that it is possible to attack DL models to which the adversary does not have access~\cite{adrel2,adrel7}. In recent years, many techniques have been proposed for increasing the robustness of DL algorithms against adversarial examples~\cite{mitrel1, mitrel2, mitrel3, mitrel4, mitrel5, mitrel6}, however, most of them have been shown to be vulnerable to future attacks~\cite{carlini2017provably}. 

Adversarial attacks in deep learning have been extensively explored for image recognition and classification applications. However, their application to the non-image domain is vastly under-explored. This also includes the lack of studies applying adversarial examples to time series analysis despite the increasing popularity of DL models in time series analysis. Very recently, the authors in~\cite{rival2019} showed that a deep neural network (DNN) univariate time series classifier (specifically ResNet~\cite{wang2017time}) are vulnerable to adversarial attacks. Unfortunately, to the knowledge of the authors, there exists no research work to date evaluating the impact of adversarial attacks on \emph{multivariate time series} (MTS) deep learning \emph{regression} models. This is indeed a major concern as potential adversarial attacks are present in many safety-critical applications that exploit DL models for time series forecasting. For instance, adding small perturbations to multivariate time series data (using false data injection methods~\cite{musleh2019survey}) that uses a DL regression model~\cite{gasparin2019deep} for smart grid electric load forecasting can generate wrong predictions, thus may lead to a nation-wide power outage.

In this paper, we apply and transfer adversarial attacks from the image domain to deep learning regression models for MTS forecasting. We present two experimental studies using two datasets from the finance and energy domain. The obtained results show that modern DL regression models are prone to adversarial attacks. We also show that adversarial time series examples crafted for one network architecture can be transferred to other architectures, thus holds their \emph{transferability} property~\cite{adrel2}. Therefore, this work highlights the importance of protecting against adversarial attacks in deep learning regression models for safety-critical MTS forecasting applications. 




To summarize, the main contributions of this paper are:

\begin{itemize}
  \item Formalize adversarial attacks on DL regression models for MTS forecasting.
  \item Crafting adversarial attacks for MTS DL regression models 
  using methods that are popular in the image domain and apply them to the finance and energy domain. To be specific, we use the fast gradient sign method (FGSM) \cite{goodfellow2014explaining} and basic iterative method (BIM) \cite{kurakin2016adversarial} to craft adversarial examples for Long Short-Term Memory (LSTM)~\cite{LSTM-org11}, Gated Recurrent Unit (GRU)~\cite{GRU-org}, and Convolutional Neural Network (CNN)~\cite{CNN-org} regression models. 
  \item An empirical study of adversarial attacks on two datasets from the finance and energy domain. We highlight the impact of such attacks in real-life scenarios using the Google stock \cite{GoogleDataset} and household electric power consumption dataset \cite{PowerDataset}.
  \item A comprehensive study of the transferability property of adversarial examples in DL regression models.
  \item A discussion on the potential defense techniques to be considered in future research on this topic.
\end{itemize}

The rest of the paper is organized as follows. Section II briefly discusses deep learning for multivariate time series regression and adversarial attacks. Section III formalizes the MTS DL regression and explains FGSM and BIM algorithm for crafting adversarial examples. Section IV compares the performance of CNN, LSTM, and GRU on Google stock household electric power consumption dataset, and evaluates the impacts of crafted adversarial examples on their performance. The transferability property of the attacks is evaluated in this section with a brief discussion on the potential defense mechanism. Section V concludes the paper.

\begin{figure}[t!]
	\centering
	\includegraphics[width=0.48\textwidth]{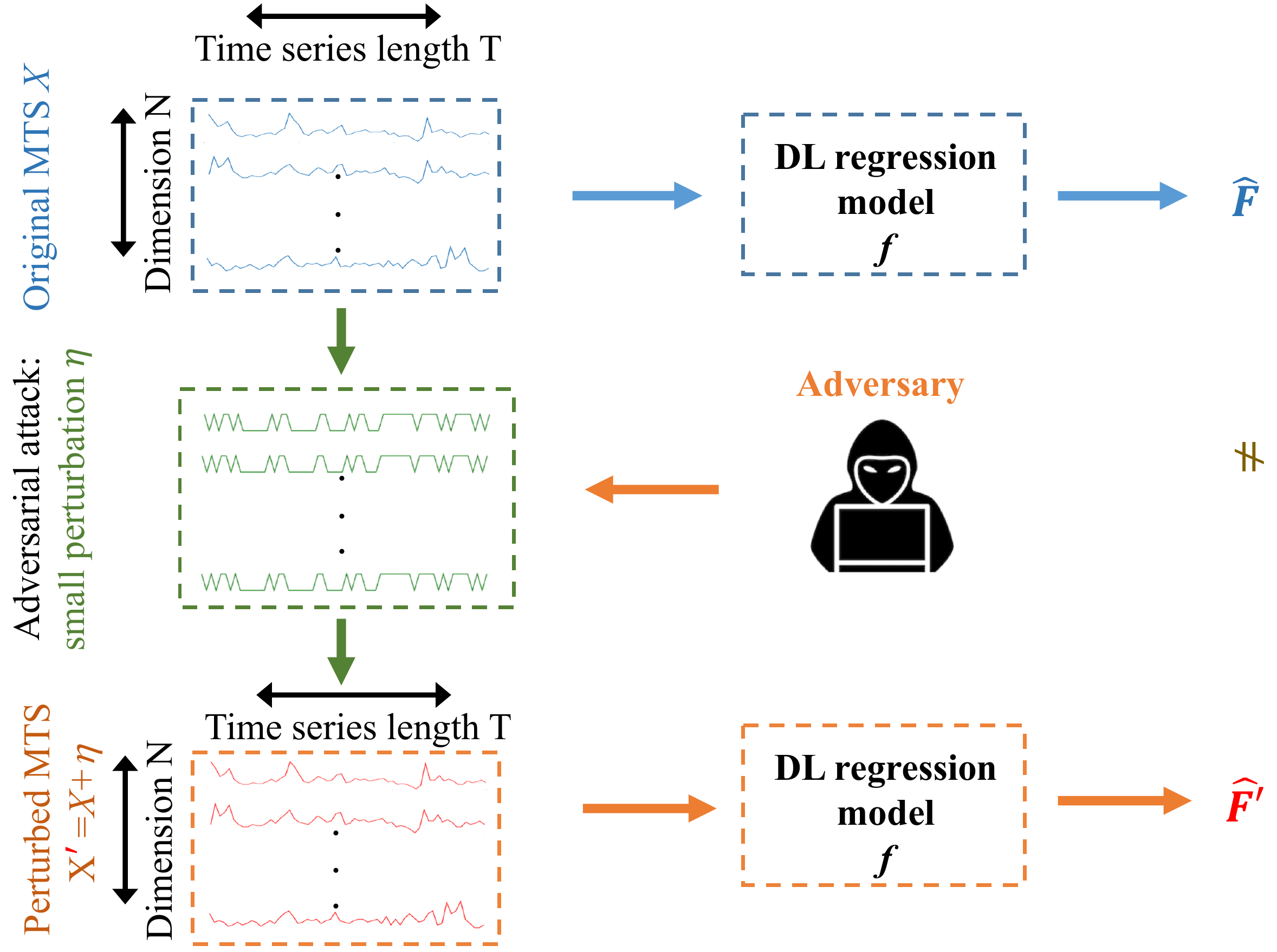}
	\caption{Example of perturbing the multivariate time series by adding imperceptible noise}
	\label{fig:FGSM_Fig}
\end{figure}

\section{Background}
\label{sec:background}


In this section, we provide an overview of DL MTS regression models and adversarial attacks in deep learning. A brief survey of state-of-the-art methods in these two areas is also presented in this section.


\subsection{Deep learning for time series forecasting}\label{DLFS}
Time series forecasting is a challenging and important problem in data science and data mining community. Therefore, hundreds of methods have been proposed for their analysis~\cite{de200625}. With the success of machine learning (ML) algorithms in different domains, ML techniques for time series forecasting is also popular~\cite{naing2015state,tealab2018time}. However, among these methods, only a few (when compared to the non-DL methods) have considered DL methods for time series forecasting\cite{borovykh2017conditional,gamboa2017deep,vengertsev2014deep}. 

In this work, we focus on the time/cost-sensitive and safety-critical applications of deep learning time series forecasting, which motivates us for investigating the impact of adversarial attacks on them. Specifically, we explore the impact of adversarial attacks on LSTM, CNN, and GRU. All of these models are known for their effectiveness in time series forecasting. LSTM is capable of learning long-term dependencies using several gates and thus suits well the time series forecasting problems. In \cite{tian2018lstm}, authors employ an LSTM model for predicting the traffic flow with missing data. The other successful applications of LSTM in time series forecasting includes petroleum production forecasting~\cite{sagheer2019time}, financial time series forecasting~\cite{cao2019financial}, solar radiation forecasting~\cite{sorkun2020time}, and remaining useful life prediction of aircraft engines~\cite{yuan2016fault}. GRU is an improvised version of Recurrent Neural Network(RNN)~\cite{RNN-org}, and also effective in time series forecasting~\cite{yamak2019comparison}. For instance, in \cite{tao2019air}, authors employ 1D convnets and bidirectional GRUs for air pollution forecasting in Beijing, China. The other applications of GRU models in time series forecasting include personalized healthcare and climate forecasting~\cite{de2019gru}, mine gas concentration forecasting~\cite{jia2020research}, smart grid bus load forecasting~\cite{shen2020short}. In \cite{dong2017cnn}, authors present a CNN-based bagging model for forecasting hourly loads in a smart grid. Apart from the energy domain, CNNs are also useful for financial time series forecasting~\cite{sezer2020financial,arratia2019convolutional}. 

In~\cite{zheng2017wide}, time-series data from smart grids are analyzed for the detection of electricity theft. In such use cases, perturbed data can help thieves to avoid being detected. Using adversarial attacks, a hacker might generate such perturbed synthetic data to bypass the system's attack detection techniques without even having access or knowledge about the DL model used for decision making~\cite{adrel2,adrel7}. Perturbing the data recorded by sensors placed on safety-critical applications (using false data injection techniques~\cite{musleh2019survey,mode2020impact}) such as aircraft engines, smart grids, gas pipeline, etc. have a catastrophic impact on human lives and productivity, whereas attacks on financial data~\cite{ngai2011application,das2012stock,akshaya2019taxonomy} has a direct impact on the economy. Indeed, the list of potential attacks presented in this section is not exhaustive due to the space limitation.

\subsection{Adversarial attacks} 
The concept of adversarial attack was proposed by \emph{Szegedy et al.} \cite{szegedy2013intriguing} at first for image recognition. The main idea is to add a small perturbation to the input images which is insignificant to human eyes, but as a result, the target model misclassifies the input images with high confidence. The severity of such attacks is shown by the researchers in a recent experiment where a strip of tape on a 35 mph limit sign was added which tricked a self-driving car into acceleration to 85 mph \cite{TeslaHack}. Based on this idea proposed in~\cite{szegedy2013intriguing}, many researchers have developed algorithms~\cite{szegedy2013intriguing, goodfellow2014explaining, kurakin2016adversarial, adrel5} for constructing such adversarial examples relying on the architecture and parameters of the DL model. Most of these adversarial attacks are proposed for image recognition tasks. A fast gradient sign method (FGSM) \cite{goodfellow2014explaining} attack was introduced in the year 2014 which signifies the presence of adversarial examples in image recognition tasks. Followed by FGSM, an iterative version of it, known as the basic iterative method (BIM) \cite{kurakin2016adversarial} was proposed in the year 2016. BIM showed more effectiveness in crafting a stealthier adversarial example, however, it comes with a higher computational cost. Comprehensive reviews of adversarial attacks in DL models in different applications can be found in~\cite{qiu2019review,xu2019adversarial,biggio2015adversarial,yuan2019adversarial}

Interestingly, the adversarial attack approaches for multivariate time series DL regression models have been ignored by the community. There are only two previous works that consider adversarial attacks on time series. In \cite{oregi2018adversarial}, the authors adopt a soft K-Nearest-Neighbours (KNN) coupled with Dynamic Time Warping (DTW) and show that the adversarial examples can fool the proposed classifier on a simulated dataset. Unfortunately, the KNN classifier is no longer considered the state-of-art classifier for time series data \cite{bagnall2017great}. The authors in \cite{rival2019}, utilize the FGSM and BIM attacks to fool Residual network (ResNet) classifiers for \emph{univariate} time series \emph{classification} tasks. In our work, we also employ the FGSM and BIM attacks, however, we apply and evaluate their impacts on DL regression models for \emph{mutivariate} time series \emph{forecasting}. 

In summary, our work sheds light on the resiliency of DL regression models for multivariate time series forecasting in real-world safety-critical and cost-critical applications (as explained in section \ref{DLFS}). This will guide the data mining, data science, and machine learning researchers to develop techniques for detecting and mitigating adversarial attacks in time series data. 

\section{adversarial examples for multivariate time series}
In this section, we formalization of the problem, and present the FGSM and BIM attack algorithms that we use to generate adversarial MTS examples for the DL models.
\subsection{Formalization of MTS regression}

\noindent \emph{Definition 1:}  Let $X$ be a multivariate time series (MTS). $X$ can be defined as a sequence such that $X = [x_1,x_2,...,x_T]$, $T=\mid X \mid$ is the length of $X$, and $x_i\in\mathbb{R}^N$ is a $N$ dimension data point at time $i\in[1,T]$.

\noindent \emph{Definition 2:} $D = {(x_1,F_1), (x_2,F_2),...,(x_T,F_T)}$ is the dataset of pairs $(x_i,F_i)$ where $F_i$ is a label corresponding to $x_i$.

\noindent \emph{Definition 3:} Time series regression task consists of training the model on $D$ in order to predict $\hat{F}$ from the possible inputs. Let $f(\cdot):\mathbb{R}^{N \times T} \rightarrow \hat{F}$ represent a DL model for regression.

\noindent \emph{Definition 4:} $J_f(\cdot,\cdot)$ denotes the cost function (e.g. mean squared error) of the model $f$.

\noindent \emph{Definition 5:} ${X}'$ denotes the adversarial example, a perturbed version of $X$ such that $\hat{F} \neq \hat{F}'$ and $\left \| X - {X}' \right \| \leq \epsilon $. where $\epsilon \geq 0 \in \mathbb{R}$ is a maximum perturbation magnitude.\\

Given a trained deep learning model $f$ and an input MTS $X$, crafting an adversarial example $X'$ can be described as a box-constrained optimization problem~\cite{yuan2019adversarial}.
\begin{gather*}
\min_{X'} {\left \| {X}'- X \right \|} ~s.t.\\
f(X^{'})=\hat{F}',~ f(X)=\hat{F}~ and~ \hat{F} \neq \hat{F}'  
\end{gather*}

Let $\eta=~X-X'$ be the perturbation added to $X$. \figurename \ref{fig:FGSM_Fig} shows the process where a perturbation $\eta$ is added to the original MTS $X$ for crafting an adversarial example $X^{'}$. 


\subsection{Fast gradient sign method}
The FGSM was first proposed in \cite{goodfellow2014explaining} where it was able to fool the GoogLeNet model by generating stealthy adversarial images. FGSM calculates the gradient of the cost function relative to the neural network input. This attack is also known as the one-shot method as the adversarial perturbation is generated by a single-step computation. Note, FGSM is an approximate solution based on linear hypothesis~\cite{qiu2019review}.
Adversarial examples are produced by the following formula:
\begin{gather}
\eta =\epsilon \cdot sign(\triangledown_x J_f(X,\hat{F})) \\
X'=X + \eta
\end{gather}

Here, $J_f$ is the cost function of model $f$, $\triangledown_x$ indicates the gradient of the model with respect to the original MTS $X$ with the correct label $\hat{F}$, $\epsilon$ denotes the hyper-parameter which controls the amplitude of the perturbation and $X'$ is adversarial MTS. Algorithm \ref{alg:fgsm} shows different steps of the FGSM attack.

\begin{algorithm}[t]
\SetAlgoLined
\Input{Original MTS $X$ and its $\hat{F}$}
\Output{Perturbed MTS $X'$}
\Parameter{$\epsilon$}
 $\eta =\epsilon \cdot sign(\triangledown_xJ_f(X,\hat{F}))$\;
  $X'=X + \eta$\;
 \caption{FGSM attack on multivariate time series} \label{alg:fgsm}
\end{algorithm}



\subsection{Basic iterative method}

The BIM \cite{kurakin2016adversarial} is an extension of FGSM. In BIM, FGSM is applied multiple times with small step size and clipping is performed after each step to ensure that they are in the range [$X-\epsilon,X+\epsilon$] i.e. $\epsilon-neighbourhood$ of the original MTS $X$. BIM is also known as Iterative-FGSM as FGSM is iterated with smaller step sizes. Algorithm \ref{alg:BIM} shows different steps of the BIM attack, where it requires three hyperparameters: the per-step small perturbation $\alpha$, the amount of maximum perturbation $\epsilon$, and the number of iterations $I$. Note, BIM does not rely on the approximation of the model, and the adversarial examples crafted through BIM are closer to the original samples when compared to FGSM. This is because the perturbations are added iteratively and hence have a better chance of fooling the network. However, compared to FGSM, BIM is computationally more expensive and slower. 

\begin{algorithm}[t]
\SetAlgoLined
\Input{Original MTS $X$ and its $\hat{F}$}
\Output{Perturbed MTS $X'$}
\Parameter{$I, \epsilon, \alpha$}
 $X' \leftarrow X$\;
 \While{$i=1 \leq I$}{
  $\eta =\alpha \cdot sign(\triangledown_xJ_f(X',\hat{F}))$\;
  $X'=X + \eta$\;
  $X'=min \{ X+\epsilon, max \{ X -\epsilon,X'\}\}$\;
  $i++$\;
 }
 \caption{BIM attack on multivariate time series} \label{alg:BIM}
 
\end{algorithm}

\begin{figure*}[ht]
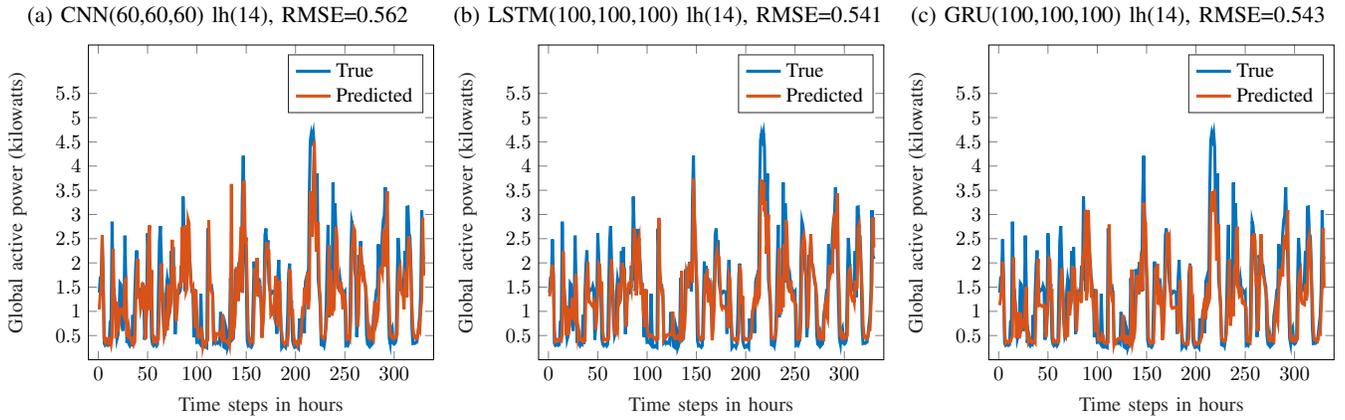

\centering
\resizebox{0.99\textwidth}{!}{
	\begin{subfigure}[t]{.33\textwidth}
		\centering
		\caption{CNN(60,60,60) lh(14), RMSE=0.562}
		{\resizebox{\textwidth}{!}{\input{Figure/Power_CNN.tikz}}\label{fig:tempSee}}
	\end{subfigure}~
	\begin{subfigure}[t]{.33\textwidth}
		\centering
		\caption{LSTM(100,100,100) lh(14), RMSE=0.541}
		{\resizebox{\textwidth}{!}{\input{Figure/Power_LSTM.tikz}}\label{fig:tempS33}}
	\end{subfigure}~
	\begin{subfigure}[t]{.33\textwidth}
		\centering
		\caption{GRU(100,100,100) lh(14), RMSE=0.543}
		{\resizebox{\textwidth}{!}{\input{Figure/Power_GRU.tikz}}\label{fig:tempS44}}
	\end{subfigure}
	}
	\caption{Comparison of deep learning algorithms for power consumption  dataset}\label{fig:Power_comp}
\end{figure*}
\section{Results}
In this section, we evaluate the crafted adversarial examples on two datasets (from the finance and energy domain) and present the obtained results. We also provide a brief discussion on potential defense mechanism for detecting the adversarial MTS examples in DL regression models. For the sake of reproducibility and to allow the research community to build on our findings, the artifacts (source code, datasets, etc.) of the following experiments are publicly available on our GitHub repository\footnote{https://github.com/dependable-cps/adversarial-MTSR}.

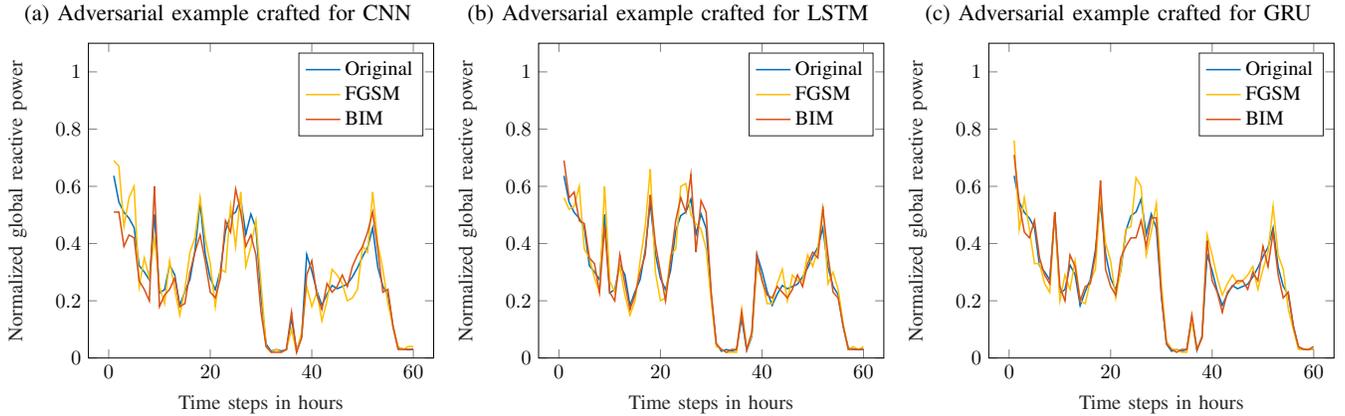
\begin{figure*}[t]
\centering
\resizebox{0.99\textwidth}{!}{
	\begin{subfigure}[t]{.33\textwidth}
		\centering
		\caption{Adversarial example crafted for CNN}
		{\resizebox{\textwidth}{!}{
%
%
\definecolor{mycolor2}{rgb}{0.00000,0.44700,0.74100}%
\definecolor{mycolor1}{rgb}{0.85000,0.32500,0.09800}%
\definecolor{mycolor3}{rgb}{0,128,0}%
\definecolor{mycolor4}{rgb}{1.0, 0.75, 0.0}%
\begin{tikzpicture}

\begin{axis}[%
width=\textwidth,
height=2.15in,
at={(2.239in,0.602in)},
legend pos=south east,
scale only axis,
xmin=-4,
xmax=64,
xlabel style={font=\color{white!15!black}},
xlabel={Time steps in hours},
ymin=0,
ymax=1.1,
xtick ={0,20,40,60},
ylabel style={font=\color{white!15!black}},
ylabel={Normalized global reactive power},
ytick ={0,0.2,0.4,0.6,0.8, 1},
axis background/.style={fill=white},
legend style={legend cell align=left, align=left,draw=white!15!black},
legend pos=north east
]
\addplot [color=mycolor2,line width = 0.75pt]
  table[row sep=crcr]{%
1	0.63681623	\\
2	0.545044952	\\
3	0.509005883	\\
4	0.488549737	\\
5	0.455597216	\\
6	0.322554584	\\
7	0.301031611	\\
8	0.273200895	\\
9	0.501108257	\\
10	0.227337697	\\
11	0.238979575	\\
12	0.324983428	\\
13	0.290953723	\\
14	0.183219746	\\
15	0.232428429	\\
16	0.274459336	\\
17	0.376765961	\\
18	0.544532253	\\
19	0.364637279	\\
20	0.278390024	\\
21	0.238756888	\\
22	0.305853047	\\
23	0.444556076	\\
24	0.497477938	\\
25	0.510020922	\\
26	0.555128019	\\
27	0.432593114	\\
28	0.502988151	\\
29	0.453158015	\\
30	0.216576211	\\
31	0.048742594	\\
32	0.023672163	\\
33	0.029410242	\\
34	0.024930604	\\
35	0.028887186	\\
36	0.140189336	\\
37	0.026334051	\\
38	0.075506484	\\
39	0.361441977	\\
40	0.304221734	\\
41	0.233873307	\\
42	0.184203712	\\
43	0.223401831	\\
44	0.253562995	\\
45	0.242066123	\\
46	0.249984464	\\
47	0.257949414	\\
48	0.283584331	\\
49	0.315526992	\\
50	0.354000083	\\
51	0.386931889	\\
52	0.454680573	\\
53	0.317831545	\\
54	0.250880391	\\
55	0.221123172	\\
56	0.110794631	\\
57	0.035583337	\\
58	0.031574968	\\
59	0.028633426	\\
60	0.031668186	\\
};
\addlegendentry{Original}

\addplot [color=mycolor4,line width=0.75pt]
  table[row sep=crcr]{%
1	0.69	\\
2	0.67	\\
3	0.46	\\
4	0.56	\\
5	0.6	\\
6	0.25	\\
7	0.35	\\
8	0.28	\\
9	0.42	\\
10	0.25	\\
11	0.19	\\
12	0.34	\\
13	0.24	\\
14	0.15	\\
15	0.24	\\
16	0.37	\\
17	0.42	\\
18	0.56	\\
19	0.42	\\
20	0.33	\\
21	0.18	\\
22	0.31	\\
23	0.3	\\
24	0.53	\\
25	0.39	\\
26	0.58	\\
27	0.32	\\
28	0.39	\\
29	0.48	\\
30	0.26	\\
31	0.04	\\
32	0.02	\\
33	0.03	\\
34	0.02	\\
35	0.03	\\
36	0.1	\\
37	0.02	\\
38	0.09	\\
39	0.25	\\
40	0.18	\\
41	0.23	\\
42	0.13	\\
43	0.2	\\
44	0.31	\\
45	0.29	\\
46	0.26	\\
47	0.2	\\
48	0.21	\\
49	0.24	\\
50	0.39	\\
51	0.37	\\
52	0.58	\\
53	0.4	\\
54	0.31	\\
55	0.19	\\
56	0.11	\\
57	0.04	\\
58	0.03	\\
59	0.04	\\
60	0.04	\\
};
\addlegendentry{FGSM}

\addplot [color=mycolor1,line width=0.75pt]
  table[row sep=crcr]{%
1	0.51	\\
2	0.51	\\
3	0.39	\\
4	0.43	\\
5	0.42	\\
6	0.27	\\
7	0.24	\\
8	0.2	\\
9	0.6	\\
10	0.18	\\
11	0.22	\\
12	0.24	\\
13	0.28	\\
14	0.18	\\
15	0.19	\\
16	0.3	\\
17	0.37	\\
18	0.43	\\
19	0.34	\\
20	0.23	\\
21	0.21	\\
22	0.28	\\
23	0.48	\\
24	0.44	\\
25	0.59	\\
26	0.51	\\
27	0.39	\\
28	0.43	\\
29	0.36	\\
30	0.16	\\
31	0.04	\\
32	0.02	\\
33	0.02	\\
34	0.02	\\
35	0.03	\\
36	0.16	\\
37	0.02	\\
38	0.07	\\
39	0.29	\\
40	0.34	\\
41	0.22	\\
42	0.17	\\
43	0.26	\\
44	0.23	\\
45	0.25	\\
46	0.29	\\
47	0.25	\\
48	0.32	\\
49	0.36	\\
50	0.39	\\
51	0.44	\\
52	0.51	\\
53	0.38	\\
54	0.23	\\
55	0.24	\\
56	0.11	\\
57	0.03	\\
58	0.03	\\
59	0.03	\\
60	0.03	\\
};
\addlegendentry{BIM}

\end{axis}
\end{tikzpicture}%
}\label{fig:attack_powerCNN}}
	\end{subfigure}~
	\begin{subfigure}[t]{.33\textwidth}
		\centering
		\caption{Adversarial example crafted for LSTM}
		{\resizebox{\textwidth}{!}{
%
%
\definecolor{mycolor2}{rgb}{0.00000,0.44700,0.74100}%
\definecolor{mycolor1}{rgb}{0.85000,0.32500,0.09800}%
\definecolor{mycolor3}{rgb}{0,128,0}%
\definecolor{mycolor4}{rgb}{1.0, 0.75, 0.0}%
\begin{tikzpicture}

\begin{axis}[%
width=\textwidth,
height=2.15in,
at={(2.239in,0.602in)},
legend pos=south east,
scale only axis,
xmin=-4,
xmax=64,
xlabel style={font=\color{white!15!black}},
xlabel={Time steps in hours},
ymin=0,
ymax=1.1,
xtick ={0,20,40,60},
ylabel style={font=\color{white!15!black}},
ylabel={Normalized global reactive power},
ytick ={0,0.2,0.4,0.6,0.8, 1},
axis background/.style={fill=white},
legend style={legend cell align=left, align=left,draw=white!15!black},
legend pos=north east
]
\addplot [color=mycolor2,line width = 0.75pt]
  table[row sep=crcr]{%
1	0.63681623	\\
2	0.545044952	\\
3	0.509005883	\\
4	0.488549737	\\
5	0.455597216	\\
6	0.322554584	\\
7	0.301031611	\\
8	0.273200895	\\
9	0.501108257	\\
10	0.227337697	\\
11	0.238979575	\\
12	0.324983428	\\
13	0.290953723	\\
14	0.183219746	\\
15	0.232428429	\\
16	0.274459336	\\
17	0.376765961	\\
18	0.544532253	\\
19	0.364637279	\\
20	0.278390024	\\
21	0.238756888	\\
22	0.305853047	\\
23	0.444556076	\\
24	0.497477938	\\
25	0.510020922	\\
26	0.555128019	\\
27	0.432593114	\\
28	0.502988151	\\
29	0.453158015	\\
30	0.216576211	\\
31	0.048742594	\\
32	0.023672163	\\
33	0.029410242	\\
34	0.024930604	\\
35	0.028887186	\\
36	0.140189336	\\
37	0.026334051	\\
38	0.075506484	\\
39	0.361441977	\\
40	0.304221734	\\
41	0.233873307	\\
42	0.184203712	\\
43	0.223401831	\\
44	0.253562995	\\
45	0.242066123	\\
46	0.249984464	\\
47	0.257949414	\\
48	0.283584331	\\
49	0.315526992	\\
50	0.354000083	\\
51	0.386931889	\\
52	0.454680573	\\
53	0.317831545	\\
54	0.250880391	\\
55	0.221123172	\\
56	0.110794631	\\
57	0.035583337	\\
58	0.031574968	\\
59	0.028633426	\\
60	0.031668186	\\
};
\addlegendentry{Original}

\addplot [color=mycolor4,line width=0.75pt]
  table[row sep=crcr]{%
1	0.56	\\
2	0.52	\\
3	0.53	\\
4	0.6	\\
5	0.38	\\
6	0.35	\\
7	0.28	\\
8	0.23	\\
9	0.6	\\
10	0.27	\\
11	0.23	\\
12	0.32	\\
13	0.22	\\
14	0.15	\\
15	0.2	\\
16	0.34	\\
17	0.46	\\
18	0.66	\\
19	0.3	\\
20	0.2	\\
21	0.21	\\
22	0.36	\\
23	0.38	\\
24	0.6	\\
25	0.61	\\
26	0.5	\\
27	0.47	\\
28	0.45	\\
29	0.38	\\
30	0.2	\\
31	0.04	\\
32	0.03	\\
33	0.02	\\
34	0.02	\\
35	0.02	\\
36	0.17	\\
37	0.03	\\
38	0.08	\\
39	0.32	\\
40	0.28	\\
41	0.19	\\
42	0.19	\\
43	0.26	\\
44	0.31	\\
45	0.2	\\
46	0.29	\\
47	0.27	\\
48	0.27	\\
49	0.36	\\
50	0.32	\\
51	0.38	\\
52	0.53	\\
53	0.26	\\
54	0.3	\\
55	0.24	\\
56	0.12	\\
57	0.03	\\
58	0.04	\\
59	0.03	\\
60	0.04	\\
};
\addlegendentry{FGSM}

\addplot [color=mycolor1,line width=0.75pt]
  table[row sep=crcr]{%
1	0.69	\\
2	0.56	\\
3	0.58	\\
4	0.48	\\
5	0.47	\\
6	0.35	\\
7	0.33	\\
8	0.23	\\
9	0.46	\\
10	0.23	\\
11	0.2	\\
12	0.36	\\
13	0.26	\\
14	0.17	\\
15	0.22	\\
16	0.3	\\
17	0.36	\\
18	0.57	\\
19	0.4	\\
20	0.3	\\
21	0.2	\\
22	0.34	\\
23	0.48	\\
24	0.56	\\
25	0.51	\\
26	0.64	\\
27	0.37	\\
28	0.55	\\
29	0.51	\\
30	0.23	\\
31	0.05	\\
32	0.03	\\
33	0.02	\\
34	0.03	\\
35	0.03	\\
36	0.16	\\
37	0.03	\\
38	0.09	\\
39	0.36	\\
40	0.27	\\
41	0.22	\\
42	0.21	\\
43	0.25	\\
44	0.23	\\
45	0.21	\\
46	0.24	\\
47	0.29	\\
48	0.25	\\
49	0.33	\\
50	0.37	\\
51	0.35	\\
52	0.52	\\
53	0.36	\\
54	0.23	\\
55	0.21	\\
56	0.11	\\
57	0.03	\\
58	0.03	\\
59	0.03	\\
60	0.03	\\
};
\addlegendentry{BIM}

\end{axis}
\end{tikzpicture}%
}\label{fig:attack_powerLSTM}}
	\end{subfigure}~
		\begin{subfigure}[t]{.33\textwidth}
		\centering
		\caption{Adversarial example crafted for GRU}
		{\resizebox{\textwidth}{!}{
%
%
\definecolor{mycolor2}{rgb}{0.00000,0.44700,0.74100}%
\definecolor{mycolor1}{rgb}{0.85000,0.32500,0.09800}%
\definecolor{mycolor3}{rgb}{0,128,0}%
\definecolor{mycolor4}{rgb}{1.0, 0.75, 0.0}%
\begin{tikzpicture}

\begin{axis}[%
width=\textwidth,
height=2.15in,
at={(2.239in,0.602in)},
legend pos=south east,
scale only axis,
xmin=-4,
xmax=64,
xlabel style={font=\color{white!15!black}},
xlabel={Time steps in hours},
ymin=0,
ymax=1.1,
xtick ={0,20,40,60},
ylabel style={font=\color{white!15!black}},
ylabel={Normalized global reactive power},
ytick ={0,0.2,0.4,0.6,0.8, 1},
axis background/.style={fill=white},
legend style={legend cell align=left, align=left,draw=white!15!black},
legend pos=north east
]
\addplot [color=mycolor2,line width = 0.75pt]
  table[row sep=crcr]{%
1	0.63681623	\\
2	0.545044952	\\
3	0.509005883	\\
4	0.488549737	\\
5	0.455597216	\\
6	0.322554584	\\
7	0.301031611	\\
8	0.273200895	\\
9	0.501108257	\\
10	0.227337697	\\
11	0.238979575	\\
12	0.324983428	\\
13	0.290953723	\\
14	0.183219746	\\
15	0.232428429	\\
16	0.274459336	\\
17	0.376765961	\\
18	0.544532253	\\
19	0.364637279	\\
20	0.278390024	\\
21	0.238756888	\\
22	0.305853047	\\
23	0.444556076	\\
24	0.497477938	\\
25	0.510020922	\\
26	0.555128019	\\
27	0.432593114	\\
28	0.502988151	\\
29	0.453158015	\\
30	0.216576211	\\
31	0.048742594	\\
32	0.023672163	\\
33	0.029410242	\\
34	0.024930604	\\
35	0.028887186	\\
36	0.140189336	\\
37	0.026334051	\\
38	0.075506484	\\
39	0.361441977	\\
40	0.304221734	\\
41	0.233873307	\\
42	0.184203712	\\
43	0.223401831	\\
44	0.253562995	\\
45	0.242066123	\\
46	0.249984464	\\
47	0.257949414	\\
48	0.283584331	\\
49	0.315526992	\\
50	0.354000083	\\
51	0.386931889	\\
52	0.454680573	\\
53	0.317831545	\\
54	0.250880391	\\
55	0.221123172	\\
56	0.110794631	\\
57	0.035583337	\\
58	0.031574968	\\
59	0.028633426	\\
60	0.031668186	\\
};
\addlegendentry{Original}

\addplot [color=mycolor4,line width=0.75pt]
  table[row sep=crcr]{%
1	0.76	\\
2	0.45	\\
3	0.56	\\
4	0.45	\\
5	0.33	\\
6	0.33	\\
7	0.26	\\
8	0.23	\\
9	0.51	\\
10	0.2	\\
11	0.29	\\
12	0.24	\\
13	0.34	\\
14	0.2	\\
15	0.19	\\
16	0.27	\\
17	0.31	\\
18	0.57	\\
19	0.4	\\
20	0.34	\\
21	0.21	\\
22	0.31	\\
23	0.45	\\
24	0.46	\\
25	0.63	\\
26	0.6	\\
27	0.36	\\
28	0.44	\\
29	0.54	\\
30	0.26	\\
31	0.06	\\
32	0.03	\\
33	0.03	\\
34	0.02	\\
35	0.02	\\
36	0.13	\\
37	0.03	\\
38	0.07	\\
39	0.43	\\
40	0.36	\\
41	0.28	\\
42	0.22	\\
43	0.26	\\
44	0.29	\\
45	0.26	\\
46	0.27	\\
47	0.29	\\
48	0.32	\\
49	0.24	\\
50	0.31	\\
51	0.37	\\
52	0.53	\\
53	0.36	\\
54	0.31	\\
55	0.17	\\
56	0.1	\\
57	0.03	\\
58	0.03	\\
59	0.03	\\
60	0.03	\\
};
\addlegendentry{FGSM}

\addplot [color=mycolor1,line width=0.75pt]
  table[row sep=crcr]{%
1	0.71	\\
2	0.54	\\
3	0.44	\\
4	0.42	\\
5	0.48	\\
6	0.35	\\
7	0.29	\\
8	0.26	\\
9	0.51	\\
10	0.25	\\
11	0.2	\\
12	0.36	\\
13	0.32	\\
14	0.2	\\
15	0.25	\\
16	0.26	\\
17	0.36	\\
18	0.62	\\
19	0.31	\\
20	0.25	\\
21	0.22	\\
22	0.35	\\
23	0.39	\\
24	0.42	\\
25	0.42	\\
26	0.48	\\
27	0.39	\\
28	0.49	\\
29	0.49	\\
30	0.22	\\
31	0.05	\\
32	0.03	\\
33	0.02	\\
34	0.03	\\
35	0.03	\\
36	0.15	\\
37	0.03	\\
38	0.07	\\
39	0.41	\\
40	0.27	\\
41	0.23	\\
42	0.16	\\
43	0.23	\\
44	0.25	\\
45	0.27	\\
46	0.27	\\
47	0.24	\\
48	0.3	\\
49	0.27	\\
50	0.39	\\
51	0.32	\\
52	0.44	\\
53	0.29	\\
54	0.21	\\
55	0.23	\\
56	0.11	\\
57	0.04	\\
58	0.03	\\
59	0.03	\\
60	0.04	\\
};
\addlegendentry{BIM}

\end{axis}
\end{tikzpicture}%
}\label{fig:attack_powerGRU}}
	\end{subfigure}
	}
	\caption{Attack signatures for power consumption dataset; FGSM ($\epsilon=0.2$) and BIM ($\alpha = 0.001$,~$\epsilon=0.2$, and~$I=200$)} \label{fig:Power_attack_sig}
\end{figure*}

\subsection{Attacks on household power consumption}
Due to the increase in demand for efficient energy needs, there is a need for a smart infrastructure to meet the growing demands and to generate energy more efficiently. Recently, deep learning \cite{kim2019predicting,wang2019data,moon2018forecasting} has shown tremendous success in forecasting the energy demands by training on the past power consumption data and forecasting the energy consumption in the future. This indeed helps in making an informed decision of how much energy should be generated for a given day in the recent future, avoids the excessive generation of surplus energy, and thus helps in reducing the loss of resources, manpower, and cost. In this context, an adversarial attack could result in incorrect predictions of global active power, which is the power consumed by electrical appliances other than the sub-metered appliances. Such an incorrect forecast may lead to either excessive surplus or inadequate generation of energy--both of which have a direct impact on cost, productivity, available resources, and environment.

In this work, we evaluate the impact of adversarial attacks on household energy forecasting using the individual household electric power consumption dataset \cite{PowerDataset}. The household power consumption dataset is a multivariate time series dataset that includes the measurements of electric power consumption in one household with a one-minute sampling rate for almost 4 years (December 2006 to November 2010) and collected via sub-meters placed in three distinct areas. The dataset is comprised of seven variables (besides the date and time) which includes global active power, global reactive power, voltage, global intensity, and sub-metering (1 to 3). We re-sample the dataset from minutes to hours and then predict global active power using seven variables or input features (global active power, global reactive power, voltage, global intensity, and sub-metering (1 to 3)). Then we use the first three years (2006 to 2009) for training our three DL models (LSTM, GRU, and CNN), and last year's data to test our models. The DL architecture of the DL models can be represented as LSTM(100,100,100) lh(14), GRU(100,100,100) lh(14), and CNN(60,60,60) lh(14). The notation LSTM(100,100,100) lh(14) refers to a network that has 100 nodes in the hidden layers of the first LSTM layer, 100 nodes in the hidden layers of the second LSTM layer, 100 nodes in the hidden layers of the third LSTM layer, and a sequence length of 14. In the end, there is a 1-dimensional output layer. In \figurename~\ref{fig:Power_comp}, we compare the performance of these three DL architectures in terms of their root mean squared error (RMSE)~\cite{chai2014root}. From \figurename~\ref{fig:Power_comp}, it is evident that the LSTM(100, 100, 100) has the best performance (with least RMSE) when predicting the global active power (without attack) which was trained with 250 epochs using Adam optimizer ~\cite{kingma2014adam} and grid search~\cite{zoller2019survey} for hyperparameter optimization to minimize the objective cost function: mean squared error (MSE). The hyperparameter settings for the evaluated DL models are shown in Table \ref{tab:Para}. 



\begin{table*}[h]
\centering
\caption{Hyperparameter settings for the DL models}

\begin{tabular}{|c|c|c|c||c|c|c|}
\hline
\multirow{2}{*}{\textbf{\makecell{DL \\ models}}} & \multicolumn{3}{c||}{\textbf{Power consumption dataset}}          & \multicolumn{3}{c|}{ \textbf{Google stock dataset}}                                      \\ \cline{2-7} 
                                    & \textbf{\makecell{Hidden \\ neurons}} & \textbf{\makecell{Batch \\ size}} & \textbf{Epochs} & \textbf{\makecell{Hidden \\ neurons}} & \textbf{\makecell{Batch \\ size}} & \textbf{Epochs} \\ \hline
CNN                                 & {60,60,60}                      & 512                 & 200                            & {60,60,60}                      & 14                  & 250                                \\ \hline
LSTM                                & {30,30,30}                      & 32                  & 250                                & {100,100,100}                     & 14                  & 300                              \\ \hline
GRU                                 & {30,30,30}                      & 32                  & 250                                & {100,100,100}                     & 14                  & 300                                \\ \hline
\end{tabular}%

\label{tab:Para}

\end{table*}

\begin{figure*}[t]
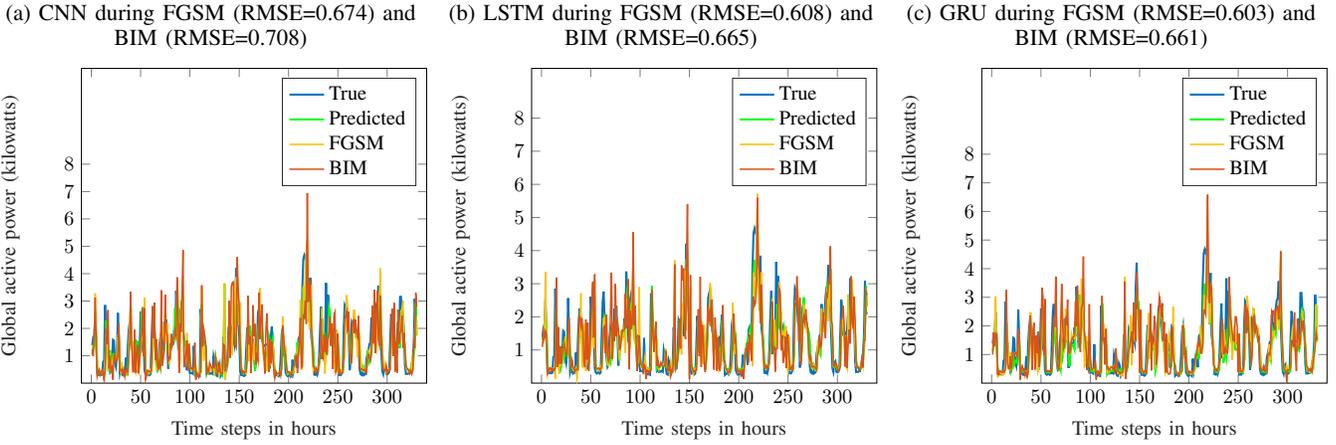

	\centering
	\captionsetup[subfigure]{justification=centering}
\resizebox{0.99\textwidth}{!}{
	\begin{subfigure}[t]{.33\textwidth}
		\centering
		\caption{CNN during FGSM (RMSE=0.674) and BIM (RMSE=0.708)}
		{\resizebox{\textwidth}{!}{\input{Figure/Power_CNN_BIM_FGSM.tikz}}\label{fig:tempPCNN}}
	\end{subfigure}~
	\begin{subfigure}[t]{.33\textwidth}
		\centering
		\caption{LSTM during FGSM (RMSE=0.608) and BIM (RMSE=0.665)}
		{\resizebox{\textwidth}{!}{\input{Figure/Power_LSTM_BIM_FGSM.tikz}}\label{fig:tempS3}}
	\end{subfigure}~
	\begin{subfigure}[t]{.33\textwidth}
		\centering
		\caption{GRU during FGSM (RMSE=0.603) and BIM (RMSE=0.661)}
		{\resizebox{\textwidth}{!}{\input{Figure/Power_GRU_BIM_FGSM.tikz}}\label{fig:tempS4}}
	\end{subfigure}
	}
	\caption{Power consumption prediction after FGSM ($\epsilon=0.2$) and BIM ($\alpha = 0.001$,~ $\epsilon=0.2$, and~$I=200$)}\label{fig:Power_comp_attack}
\end{figure*}

\figurename~\ref{fig:Power_attack_sig} shows an example of the normalized FGSM and BIM attack signatures (adversarial examples) generated for the global reactive power variable (an input feature in the form of a time series). Similar adversarial examples are generated for the remaining five input features to evaluate their impact on the LSTM, GRU and CNN models for energy consumption prediction (global active power prediction). As shown in \figurename~\ref{fig:Power_attack_sig}, the adversarial attack generated using BIM is close to the original time series data which makes such attack stealthy, hard to detect and often bypass the attack detection algorithms. The impact of the generated adversarial examples on the household electric power consumption dataset is shown in \figurename~\ref{fig:Power_comp_attack}. For the FGSM attack (with $\epsilon=0.2$), we observe that the RMSE for the CNN, LSTM and GRU model (under attack) are increased by 19.9\%, 12.3\%, and 11\%, respectively, when compared to the models without attack. For the BIM attack (with $\alpha=0.001$, $\epsilon=0.2$, and $I=200$), we also observe the similar trend, that is the RMSE of the CNN, LSTM and GRU models increased in a similar fashion, specifically by 25.9\%, 22.9\%, and 21.7\%, respectively for the household electric power consumption dataset. We observe that for both FGSM and BIM attacks, it is evident that the CNN model is more sensitive to adversarial attacks when compared to the other DL models. Also, BIM results in a larger RMSE when compared to the FGSM. This means BIM is not only stealthier that FGSM, but also has a stronger impact on DL regression models for the this dataset. 

For instance, as shown in \figurename~\ref{fig:tempPCNN}, the CNN MTS regression model forecasts the global active power (without attack) to be 2.10 kW and 4.51 kW on 161st hour and 219th hour, respectively. After performing the FGSM and BIM attack, the same CNN MTS regression model forecasts the global active power to be 1.36 kW and 0.37 kW on 161st hour, and 5.24 kW and 6.94 kW on 219th hour, respectively. This represents a 35.2\% and 82.3\% decrease, and a 16\% and 53.8\% increase in the predicted values on the 161st and 219th hour respectively (when compared to the without attack situation). Such an under-prediction as a consequence of attack may result in the inadequate generation of energy, thus leading to a failure of meeting the future energy demands with a potential power outage. In contrast, over-prediction may result in the surplus generation of energy leading to increased cost and waste of resources.

\begin{figure*}[t]
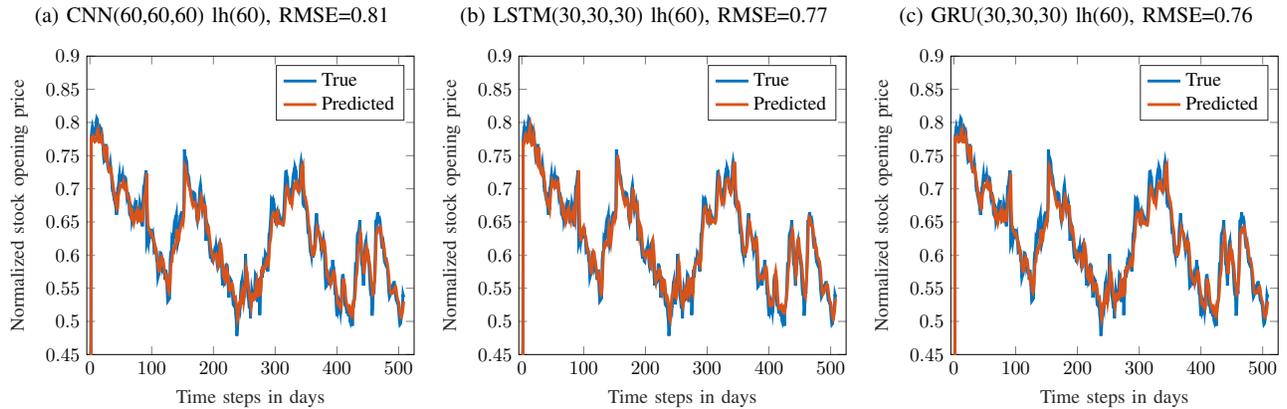

	\centering
	\captionsetup[subfigure]{justification=centering}
\resizebox{0.95\textwidth}{!}{
	\begin{subfigure}[t]{.33\textwidth}
		\centering
		\caption{CNN(60,60,60) lh(60), RMSE=0.81}
		{\resizebox{\textwidth}{!}{\input{Figure/Google_CNN.tikz}}\label{fig:tempSe1}}
	\end{subfigure}~
	\begin{subfigure}[t]{.33\textwidth}
		\centering
		\caption{LSTM(30,30,30) lh(60), RMSE=0.77}
		{\resizebox{\textwidth}{!}{\input{Figure/Google_LSTM.tikz}}\label{fig:tempS31}}
	\end{subfigure}~
	\begin{subfigure}[t]{.33\textwidth}
		\centering
		\caption{GRU(30,30,30) lh(60), RMSE=0.76}
		{\resizebox{\textwidth}{!}{\input{Figure/Google_GRU.tikz}}\label{fig:tempS41}}
	\end{subfigure}
	}
	\caption{Comparison of deep learning algorithms for Google stock dataset}\label{fig:Google_comp}
\end{figure*}


\subsection{Attacks on stock prices}
Data scientists and financial theorists have been employed for the past 50 years to make sense of the market by increasing the return on the investment. However, due to the multidimensional nature, the scale of the problem, and its inherent variation with time makes it an overwhelming task. Advancements in DL algorithms and their application to finance \cite{sezer2020financial,gan2020machine,sirignano2019universal,lee2019multimodal} has shown tremendous prospect to revolutionize this domain including stock market analysis and prediction. DL algorithms can learn the multivariate nature of the stocks and can make more accurate predictions~\cite{long2019deep,song2019study}. In this context, an adversarial attack could result in incorrect stock price predictions, which may, in turn, result in a diminishing return of the investment, and have a significant impact on the stock market.

In this work, we evaluate the impact of adversarial attacks on Google stock prediction using the Google stock dataset \cite{GoogleDataset}. The Google stock dataset contains Google stock prices for the past 5 years. This multivariate time series dataset has six variables namely date, close, open, volume, high, and low. We use 30\% of the latest stock data as our test dataset and we train our three DL models (LSTM, GRU, and CNN) on the remaining 70\% of the data. To predict the Google stock prices, we consider the average stock prices and volume of the stocks traded from the previous days as input features. As the Google stock price prediction is dependant on multiple input features, it is a multivariate regression problem. We utilize the past 60 days of data to predict the stock price of the next day. The architectures of our DL models can be represented as LSTM(30,30,30) lh(60), GRU(30,30,30) lh(60), and CNN(60,60,60) lh(60). From \figurename~\ref{fig:Google_comp}, it is evident that the GRU(30, 30, 30) has the best performance (with least RMSE) when predicting stock opening prices (without attack) which was trained with 300 epochs using Adam optimizer~\cite{kingma2014adam} and grid search~\cite{zoller2019survey} for hyperparameter optimization to minimize the objective cost function: mean squared error (MSE). The hyperparameter settings for the evaluated DL models are shown in Table \ref{tab:Para}.


\begin{figure*}[t]
\centering
\captionsetup[subfigure]{justification=centering}
\resizebox{0.99\textwidth}{!}{
	\begin{subfigure}[t]{.33\textwidth}
		\centering
		\caption{Adversarial example crafted for CNN}
		{\resizebox{\textwidth}{!}{
%
%
\definecolor{mycolor2}{rgb}{0.00000,0.44700,0.74100}%
\definecolor{mycolor1}{rgb}{0.85000,0.32500,0.09800}%
\definecolor{mycolor3}{rgb}{0,128,0}%
\definecolor{mycolor4}{rgb}{1.0, 0.75, 0.0}%

\begin{tikzpicture}

\begin{axis}[%
width=\textwidth,
height=2.15in,
at={(2.239in,0.602in)},
legend pos=south east,
scale only axis,
xmin=-4,
xmax=64,
xlabel style={font=\color{white!15!black}},
xlabel={Time steps in days},
ymin=0,
ymax=0.5,
xtick ={0,20,40,60},
ylabel style={font=\color{white!15!black}},
ylabel={Normalized volume of the stock},
ytick ={0,0.05,0.1,0.15,0.2,0.25,0.3,0.35,0.4,0.45},
scaled y ticks = false, 
scaled x ticks = false, 
y tick label style={/pgf/number format/.cd, fixed, fixed zerofill,precision=2},
axis background/.style={fill=white},
legend style={legend cell align=left, align=left,draw=white!15!black},
legend pos=north east
]
\addplot [color=mycolor2,line width = 0.75pt]
  table[row sep=crcr]{%
1	0.339389253	\\
2	0.266506651	\\
3	0.19703841	\\
4	0.221643796	\\
5	0.256525374	\\
6	0.154710913	\\
7	0.097672905	\\
8	0.084458325	\\
9	0.099878983	\\
10	0.106760284	\\
11	0.082705341	\\
12	0.104044316	\\
13	0.119931169	\\
14	0.126682104	\\
15	0.104466457	\\
16	0.149965619	\\
17	0.177489833	\\
18	0.35215989	\\
19	0.27340769	\\
20	0.216165026	\\
21	0.119463539	\\
22	0.096068231	\\
23	0.140817408	\\
24	0.156768053	\\
25	0.159409731	\\
26	0.120534189	\\
27	0.143816268	\\
28	0.182031817	\\
29	0.214280959	\\
30	0.104593682	\\
31	0.114373085	\\
32	0.13929492	\\
33	0.147641646	\\
34	0.162722438	\\
35	0.134109989	\\
36	0.136488782	\\
37	0.134920087	\\
38	0.154789509	\\
39	0.105788597	\\
40	0.125502711	\\
41	0.085642652	\\
42	0.093614879	\\
43	0.092483936	\\
44	0.059200392	\\
45	0.030468254	\\
46	0.078530587	\\
47	0.296886888	\\
48	0.131189366	\\
49	0.147963478	\\
50	0.165745434	\\
51	0.124670093	\\
52	0.13835957	\\
53	0.114286683	\\
54	0.075623243	\\
55	0.097502523	\\
56	0.120932643	\\
57	0.117318202	\\
58	0.108104495	\\
59	0.137290266	\\
60	0.113113571	\\
};
\addlegendentry{Original}

\addplot [color=mycolor4,line width=0.75pt]
  table[row sep=crcr]{%
1	0.29	\\
2	0.22	\\
3	0.17	\\
4	0.14	\\
5	0.31	\\
6	0.13	\\
7	0.11	\\
8	0.11	\\
9	0.09	\\
10	0.11	\\
11	0.08	\\
12	0.09	\\
13	0.09	\\
14	0.14	\\
15	0.09	\\
16	0.16	\\
17	0.13	\\
18	0.26	\\
19	0.32	\\
20	0.19	\\
21	0.11	\\
22	0.09	\\
23	0.1	\\
24	0.18	\\
25	0.18	\\
26	0.15	\\
27	0.09	\\
28	0.16	\\
29	0.21	\\
30	0.09	\\
31	0.11	\\
32	0.12	\\
33	0.14	\\
34	0.13	\\
35	0.13	\\
36	0.08	\\
37	0.16	\\
38	0.2	\\
39	0.13	\\
40	0.16	\\
41	0.09	\\
42	0.09	\\
43	0.1	\\
44	0.05	\\
45	0.03	\\
46	0.06	\\
47	0.19	\\
48	0.08	\\
49	0.18	\\
50	0.21	\\
51	0.1	\\
52	0.18	\\
53	0.11	\\
54	0.07	\\
55	0.07	\\
56	0.12	\\
57	0.12	\\
58	0.09	\\
59	0.11	\\
60	0.11	\\
};
\addlegendentry{FGSM}

\addplot [color=mycolor1,line width=0.75pt]
  table[row sep=crcr]{%
1	0.29	\\
2	0.27	\\
3	0.23	\\
4	0.24	\\
5	0.29	\\
6	0.17	\\
7	0.1	\\
8	0.08	\\
9	0.1	\\
10	0.09	\\
11	0.08	\\
12	0.08	\\
13	0.11	\\
14	0.1	\\
15	0.11	\\
16	0.13	\\
17	0.19	\\
18	0.31	\\
19	0.24	\\
20	0.25	\\
21	0.13	\\
22	0.11	\\
23	0.13	\\
24	0.13	\\
25	0.17	\\
26	0.1	\\
27	0.13	\\
28	0.22	\\
29	0.2	\\
30	0.11	\\
31	0.09	\\
32	0.17	\\
33	0.15	\\
34	0.16	\\
35	0.13	\\
36	0.12	\\
37	0.14	\\
38	0.13	\\
39	0.1	\\
40	0.15	\\
41	0.1	\\
42	0.08	\\
43	0.09	\\
44	0.07	\\
45	0.03	\\
46	0.07	\\
47	0.33	\\
48	0.11	\\
49	0.16	\\
50	0.13	\\
51	0.13	\\
52	0.12	\\
53	0.13	\\
54	0.08	\\
55	0.11	\\
56	0.11	\\
57	0.1	\\
58	0.1	\\
59	0.14	\\
60	0.1	\\
};
\addlegendentry{BIM}

\end{axis}
\end{tikzpicture}%
}\label{fig:attack_googleCNN}}
	\end{subfigure}~
	\begin{subfigure}[t]{.33\textwidth}
		\centering
		\caption{Adversarial example crafted for LSTM}
		{\resizebox{\textwidth}{!}{
%
%
\definecolor{mycolor2}{rgb}{0.00000,0.44700,0.74100}%
\definecolor{mycolor1}{rgb}{0.85000,0.32500,0.09800}%
\definecolor{mycolor3}{rgb}{0,128,0}%
\definecolor{mycolor4}{rgb}{1.0, 0.75, 0.0}%
\begin{tikzpicture}

\begin{axis}[%
width=\textwidth,
height=2.15in,
at={(2.239in,0.602in)},
legend pos=south east,
scale only axis,
xmin=-4,
xmax=64,
xlabel style={font=\color{white!15!black}},
xlabel={Time steps in days},
ymin=0,
ymax=0.5,
xtick ={0,20,40,60},
ylabel style={font=\color{white!15!black}},
ylabel={Normalized volume of the stock},
ytick ={0,0.05,0.1,0.15,0.2,0.25,0.3,0.35,0.4,0.45},
scaled y ticks = false, 
scaled x ticks = false, 
y tick label style={/pgf/number format/.cd, fixed, fixed zerofill,precision=2},
axis background/.style={fill=white},
legend style={legend cell align=left, align=left,draw=white!15!black},
legend pos=north east
]
\addplot [color=mycolor2,line width = 0.75pt]
  table[row sep=crcr]{%
1	0.339389253	\\
2	0.266506651	\\
3	0.19703841	\\
4	0.221643796	\\
5	0.256525374	\\
6	0.154710913	\\
7	0.097672905	\\
8	0.084458325	\\
9	0.099878983	\\
10	0.106760284	\\
11	0.082705341	\\
12	0.104044316	\\
13	0.119931169	\\
14	0.126682104	\\
15	0.104466457	\\
16	0.149965619	\\
17	0.177489833	\\
18	0.35215989	\\
19	0.27340769	\\
20	0.216165026	\\
21	0.119463539	\\
22	0.096068231	\\
23	0.140817408	\\
24	0.156768053	\\
25	0.159409731	\\
26	0.120534189	\\
27	0.143816268	\\
28	0.182031817	\\
29	0.214280959	\\
30	0.104593682	\\
31	0.114373085	\\
32	0.13929492	\\
33	0.147641646	\\
34	0.162722438	\\
35	0.134109989	\\
36	0.136488782	\\
37	0.134920087	\\
38	0.154789509	\\
39	0.105788597	\\
40	0.125502711	\\
41	0.085642652	\\
42	0.093614879	\\
43	0.092483936	\\
44	0.059200392	\\
45	0.030468254	\\
46	0.078530587	\\
47	0.296886888	\\
48	0.131189366	\\
49	0.147963478	\\
50	0.165745434	\\
51	0.124670093	\\
52	0.13835957	\\
53	0.114286683	\\
54	0.075623243	\\
55	0.097502523	\\
56	0.120932643	\\
57	0.117318202	\\
58	0.108104495	\\
59	0.137290266	\\
60	0.113113571	\\
};
\addlegendentry{Original}

\addplot [color=mycolor4,line width=0.75pt]
  table[row sep=crcr]{%
1	0.42	\\
2	0.23	\\
3	0.24	\\
4	0.27	\\
5	0.25	\\
6	0.19	\\
7	0.08	\\
8	0.07	\\
9	0.09	\\
10	0.13	\\
11	0.07	\\
12	0.1	\\
13	0.09	\\
14	0.11	\\
15	0.08	\\
16	0.15	\\
17	0.18	\\
18	0.28	\\
19	0.33	\\
20	0.26	\\
21	0.14	\\
22	0.1	\\
23	0.11	\\
24	0.13	\\
25	0.2	\\
26	0.12	\\
27	0.14	\\
28	0.19	\\
29	0.19	\\
30	0.1	\\
31	0.13	\\
32	0.14	\\
33	0.18	\\
34	0.12	\\
35	0.16	\\
36	0.16	\\
37	0.14	\\
38	0.18	\\
39	0.1	\\
40	0.13	\\
41	0.06	\\
42	0.07	\\
43	0.11	\\
44	0.05	\\
45	0.03	\\
46	0.08	\\
47	0.28	\\
48	0.1	\\
49	0.15	\\
50	0.12	\\
51	0.14	\\
52	0.12	\\
53	0.1	\\
54	0.07	\\
55	0.08	\\
56	0.11	\\
57	0.14	\\
58	0.13	\\
59	0.1	\\
60	0.13	\\
};
\addlegendentry{FGSM}

\addplot [color=mycolor1,line width=0.75pt]
  table[row sep=crcr]{%
1	0.37	\\
2	0.28	\\
3	0.17	\\
4	0.25	\\
5	0.22	\\
6	0.14	\\
7	0.1	\\
8	0.08	\\
9	0.1	\\
10	0.1	\\
11	0.07	\\
12	0.1	\\
13	0.12	\\
14	0.13	\\
15	0.1	\\
16	0.14	\\
17	0.16	\\
18	0.35	\\
19	0.28	\\
20	0.18	\\
21	0.13	\\
22	0.08	\\
23	0.12	\\
24	0.14	\\
25	0.17	\\
26	0.13	\\
27	0.15	\\
28	0.17	\\
29	0.23	\\
30	0.09	\\
31	0.12	\\
32	0.13	\\
33	0.14	\\
34	0.14	\\
35	0.12	\\
36	0.12	\\
37	0.12	\\
38	0.16	\\
39	0.1	\\
40	0.13	\\
41	0.1	\\
42	0.08	\\
43	0.09	\\
44	0.06	\\
45	0.03	\\
46	0.07	\\
47	0.28	\\
48	0.11	\\
49	0.13	\\
50	0.16	\\
51	0.13	\\
52	0.15	\\
53	0.1	\\
54	0.08	\\
55	0.09	\\
56	0.13	\\
57	0.1	\\
58	0.1	\\
59	0.13	\\
60	0.11	\\
};
\addlegendentry{BIM}

\end{axis}
\end{tikzpicture}%
}\label{fig:attack_googleLSTM}}
	\end{subfigure}~
		\begin{subfigure}[t]{.33\textwidth}
		\centering
		\caption{Adversarial example crafted for GRU}
		{\resizebox{\textwidth}{!}{
%
%
\definecolor{mycolor2}{rgb}{0.00000,0.44700,0.74100}%
\definecolor{mycolor1}{rgb}{0.85000,0.32500,0.09800}%
\definecolor{mycolor3}{rgb}{0,128,0}%
\definecolor{mycolor4}{rgb}{1.0, 0.75, 0.0}%
\begin{tikzpicture}

\begin{axis}[%
width=\textwidth,
height=2.15in,
at={(2.239in,0.602in)},
legend pos=south east,
scale only axis,
xmin=-4,
xmax=64,
xlabel style={font=\color{white!15!black}},
xlabel={Time steps in days},
ymin=0,
ymax=0.5,
xtick ={0,20,40,60},
ylabel style={font=\color{white!15!black}},
ylabel={Normalized volume of the stock},
ytick ={0,0.05,0.1,0.15,0.2,0.25,0.3,0.35,0.4,0.45},
scaled y ticks = false, 
scaled x ticks = false, 
y tick label style={/pgf/number format/.cd, fixed, fixed zerofill,precision=2},
axis background/.style={fill=white},
legend style={legend cell align=left, align=left,draw=white!15!black},
legend pos=north east
]
\addplot [color=mycolor2,line width = 0.75pt]
  table[row sep=crcr]{%
1	0.339389253	\\
2	0.266506651	\\
3	0.19703841	\\
4	0.221643796	\\
5	0.256525374	\\
6	0.154710913	\\
7	0.097672905	\\
8	0.084458325	\\
9	0.099878983	\\
10	0.106760284	\\
11	0.082705341	\\
12	0.104044316	\\
13	0.119931169	\\
14	0.126682104	\\
15	0.104466457	\\
16	0.149965619	\\
17	0.177489833	\\
18	0.35215989	\\
19	0.27340769	\\
20	0.216165026	\\
21	0.119463539	\\
22	0.096068231	\\
23	0.140817408	\\
24	0.156768053	\\
25	0.159409731	\\
26	0.120534189	\\
27	0.143816268	\\
28	0.182031817	\\
29	0.214280959	\\
30	0.104593682	\\
31	0.114373085	\\
32	0.13929492	\\
33	0.147641646	\\
34	0.162722438	\\
35	0.134109989	\\
36	0.136488782	\\
37	0.134920087	\\
38	0.154789509	\\
39	0.105788597	\\
40	0.125502711	\\
41	0.085642652	\\
42	0.093614879	\\
43	0.092483936	\\
44	0.059200392	\\
45	0.030468254	\\
46	0.078530587	\\
47	0.296886888	\\
48	0.131189366	\\
49	0.147963478	\\
50	0.165745434	\\
51	0.124670093	\\
52	0.13835957	\\
53	0.114286683	\\
54	0.075623243	\\
55	0.097502523	\\
56	0.120932643	\\
57	0.117318202	\\
58	0.108104495	\\
59	0.137290266	\\
60	0.113113571	\\
};
\addlegendentry{Original}

\addplot [color=mycolor4,line width=0.75pt]
  table[row sep=crcr]{%
1	0.42	\\
2	0.25	\\
3	0.15	\\
4	0.2	\\
5	0.3	\\
6	0.16	\\
7	0.1	\\
8	0.06	\\
9	0.12	\\
10	0.08	\\
11	0.07	\\
12	0.11	\\
13	0.14	\\
14	0.13	\\
15	0.1	\\
16	0.18	\\
17	0.17	\\
18	0.4	\\
19	0.29	\\
20	0.2	\\
21	0.13	\\
22	0.11	\\
23	0.13	\\
24	0.17	\\
25	0.13	\\
26	0.09	\\
27	0.12	\\
28	0.19	\\
29	0.2	\\
30	0.13	\\
31	0.09	\\
32	0.11	\\
33	0.12	\\
34	0.18	\\
35	0.12	\\
36	0.17	\\
37	0.14	\\
38	0.17	\\
39	0.12	\\
40	0.15	\\
41	0.09	\\
42	0.11	\\
43	0.09	\\
44	0.06	\\
45	0.03	\\
46	0.07	\\
47	0.27	\\
48	0.16	\\
49	0.18	\\
50	0.18	\\
51	0.12	\\
52	0.12	\\
53	0.12	\\
54	0.06	\\
55	0.08	\\
56	0.09	\\
57	0.11	\\
58	0.12	\\
59	0.16	\\
60	0.1	\\
};
\addlegendentry{FGSM}

\addplot [color=mycolor1,line width=0.75pt]
  table[row sep=crcr]{%
1	0.32	\\
2	0.27	\\
3	0.18	\\
4	0.24	\\
5	0.23	\\
6	0.16	\\
7	0.1	\\
8	0.08	\\
9	0.09	\\
10	0.12	\\
11	0.07	\\
12	0.09	\\
13	0.12	\\
14	0.13	\\
15	0.1	\\
16	0.15	\\
17	0.17	\\
18	0.34	\\
19	0.24	\\
20	0.19	\\
21	0.11	\\
22	0.1	\\
23	0.12	\\
24	0.15	\\
25	0.16	\\
26	0.1	\\
27	0.12	\\
28	0.17	\\
29	0.2	\\
30	0.1	\\
31	0.12	\\
32	0.14	\\
33	0.14	\\
34	0.17	\\
35	0.12	\\
36	0.14	\\
37	0.14	\\
38	0.17	\\
39	0.11	\\
40	0.12	\\
41	0.09	\\
42	0.08	\\
43	0.09	\\
44	0.06	\\
45	0.03	\\
46	0.06	\\
47	0.29	\\
48	0.14	\\
49	0.15	\\
50	0.17	\\
51	0.12	\\
52	0.14	\\
53	0.11	\\
54	0.07	\\
55	0.1	\\
56	0.13	\\
57	0.12	\\
58	0.11	\\
59	0.14	\\
60	0.11	\\
};
\addlegendentry{BIM}

\end{axis}
\end{tikzpicture}%
}\label{fig:attack_googleGRU}}
	\end{subfigure}
	}
	\caption{Attack signatures for google stock dataset; FGSM ($\epsilon=0.2$) and BIM ($\alpha = 0.001$,~$\epsilon=0.2$, and~$I=200$)}\label{fig:Google_attack_sig}
\end{figure*}
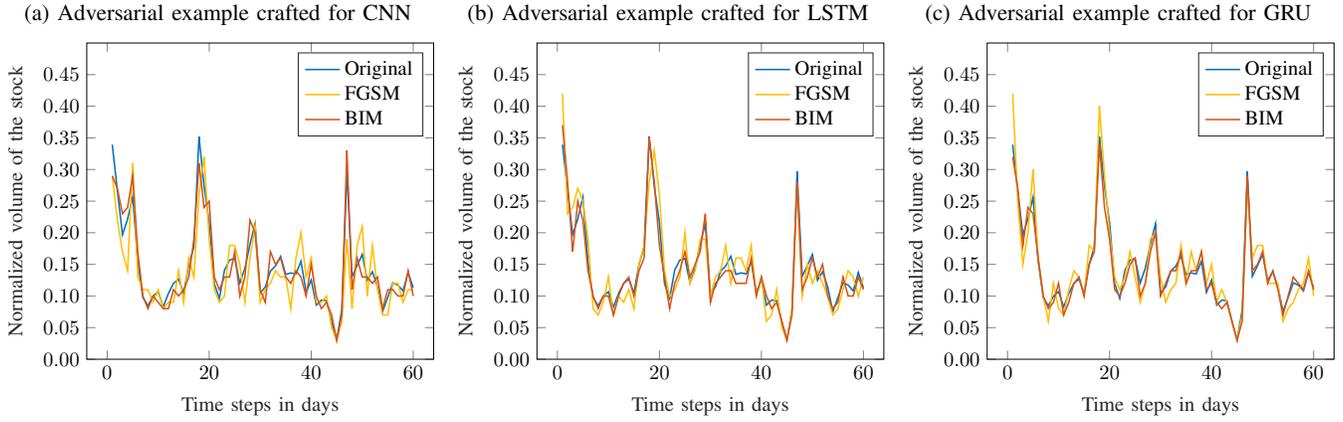

\begin{figure*}[t]
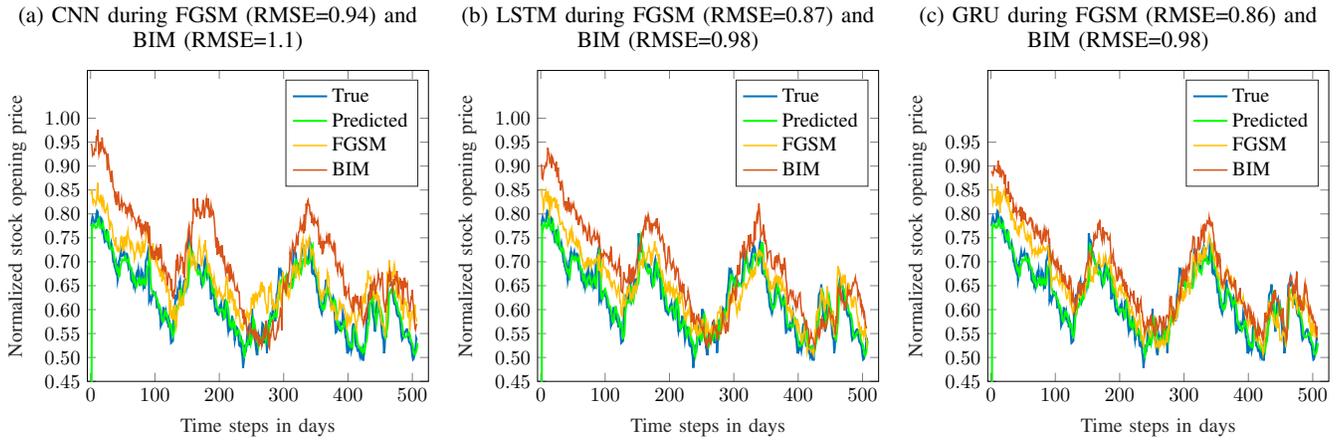

	\centering
	\captionsetup[subfigure]{justification=centering}
\resizebox{0.99\textwidth}{!}{
	\begin{subfigure}[t]{.33\textwidth}
		\centering
		\caption{CNN during FGSM (RMSE=0.94) and BIM (RMSE=1.1)}
		{\resizebox{\textwidth}{!}{\input{Figure/Google_CNN_FGSM_BIM.tikz}}\label{fig:GooglecNn}}
	\end{subfigure}~
	\begin{subfigure}[t]{.33\textwidth}
		\centering
		\caption{LSTM during FGSM (RMSE=0.87) and BIM (RMSE=0.98)}
		{\resizebox{\textwidth}{!}{\input{Figure/Google_LSTM_FGSM_BIM.tikz}}\label{fig:tempS333}}
	\end{subfigure}~
	\begin{subfigure}[t]{.33\textwidth}
		\centering
		\caption{GRU during FGSM (RMSE=0.86) and BIM (RMSE=0.98)}
		{\resizebox{\textwidth}{!}{\input{Figure/Google_GRU_FGSM_BIM.tikz}}\label{fig:tempS444}}
	\end{subfigure}
	}
	\caption{Stock price prediction after FGSM ($\epsilon=0.2$) and BIM ( $\alpha = 0.001$,~ $\epsilon=0.2$, and~$I=200$ )}\label{fig:Google_comp_attack}
\end{figure*}

\figurename~\ref{fig:Google_attack_sig} shows an example of the normalized  FGSM and BIM attack signatures (adversarial examples) generated for the volume of stocks traded (an input feature in the form of a time series). Similar adversarial examples are also generated for other input features to evaluate their impact on the LSTM, GRU and CNN models for the Google stock prediction (stock opening price). From~\figurename \ref{fig:Google_attack_sig}, we observe that the adversarial attack generated using BIM is close to the original time series data, which makes such attacks hard to detect and thus have high chances of bypassing the attack detection methods. The impact of the crafted adversarial examples on the Google stock dataset is shown in \figurename~\ref{fig:Google_comp_attack}. For the FGSM attack (with $\epsilon=0.2$), we observe that the RMSE for the CNN, LSTM and GRU model (under attack) are increased by 16\%, 12.9\%, and 13.1\%, respectively, when compared to the models without attack. For the BIM attack (with $\alpha = 0.001$, $\epsilon = 0.2$ and $I= 200$), we also observe the similar trend, that is the RMSE for the CNN, LSTM and GRU model (under attack) are increased by 35.2\%, 27.2\% and 28.9\%, respectively. Similar to our observation on the power consumption dataset, we notice that the CNN model is more sensitive to adversarial attacks when compared to the other DL models. Moreover, we also observe that BIM results in a larger RMSE when compared to the FGSM.

For instance, as shown in \figurename~\ref{fig:GooglecNn}, the CNN MTS regression model forecasts the normalized stock opening price (without attack) to be \$0.781 on day 11 and \$0.662 on day 297. After performing the FGSM and BIM attack, the same CNN MTS regression model forecasts the normalized stock opening price to be \$0.864 and \$0.975 on day 11, and \$0.607 and \$0.556 on day 297, respectively. This represents a 10.6\% and 24.8\% increase, and an 8.3\% and 16\% decrease in the predicted stock prices on day 11 and 297, respectively (when compared to the without attack situation). Such an over-prediction and under-prediction in stock prices may result in investors investing more and investing less in a particular stock whereas the stock prices are decreasing and increasing, respectively, thus leading to a loss in the return of investment in both cases.

\subsection{Performance variation vs. the amount of perturbation}
\begin{figure}[ht]
	\centering
	\begin{subfigure}[t]{.24\textwidth}
		\centering
		\caption{Power consumption dataset (LSTM model)}
		{\resizebox{\textwidth}{!}{
%
%
\definecolor{mycolor2}{rgb}{0.00000,0.44700,0.74100}%
\definecolor{mycolor1}{rgb}{0.85000,0.32500,0.09800}%
\definecolor{mycolor3}{rgb}{0,128,0}%
\definecolor{mycolor4}{rgb}{1.0, 0.75, 0.0}%
\begin{tikzpicture}

\begin{axis}[%
width=\textwidth,
height=2.15in,
at={(2.239in,0.602in)},
legend pos=south east,
scale only axis,
xmin=0,
xmax=1.5,
xlabel style={font=\color{white!15!black}},
xlabel={Amount of perturbation ($\epsilon$)},
ymin=0,
ymax=1.4,
xtick ={0,0.2,0.4,0.6,0.8,1,1.2,1.4},
ylabel style={font=\color{white!15!black}},
ylabel={RMSE},
ytick ={0,0.1,0.2,0.3,0.4,0.5,0.6,0.7,0.8,0.9,1},
axis background/.style={fill=white},
legend style={legend cell align=left, align=left,draw=white!15!black},
legend pos=north east
]

\addplot [color=mycolor4,line width=0.75pt]
  table[row sep=crcr]{%
0	0	\\
0.2	0.608	\\
0.4	0.623	\\
0.6	0.667	\\
0.8	0.685	\\
1.0	0.714	\\
1.2	0.732	\\
1.4	0.744	\\
};
\addlegendentry{FGSM}

\addplot [color=mycolor1,line width=0.75pt]
  table[row sep=crcr]{%
0	0	\\
0.2	0.665	\\
0.4	0.693	\\
0.6	0.746	\\
0.8	0.781	\\
1.0	0.880	\\
1.2	0.920	\\
1.4	0.940	\\
};
\addlegendentry{BIM}

\end{axis}
\end{tikzpicture}%
}\label{fig:PowerEps}}
	\end{subfigure}~
	\begin{subfigure}[t]{.24\textwidth}
		\centering
		\caption{Google stock dataset (GRU model)}
		{\resizebox{\textwidth}{!}{
%
%
\definecolor{mycolor2}{rgb}{0.00000,0.44700,0.74100}%
\definecolor{mycolor1}{rgb}{0.85000,0.32500,0.09800}%
\definecolor{mycolor3}{rgb}{0,128,0}%
\definecolor{mycolor4}{rgb}{1.0, 0.75, 0.0}%
\begin{tikzpicture}

\begin{axis}[%
width=\textwidth,
height=2.15in,
at={(2.239in,0.602in)},
legend pos=south east,
scale only axis,
xmin=0,
xmax=1.5,
xlabel style={font=\color{white!15!black}},
xlabel={Amount of perturbation ($\epsilon$)},
ymin=0,
ymax=2.5,
xtick ={0,0.2,0.4,0.6,0.8,1,1.2,1.4},
ylabel style={font=\color{white!15!black}},
ylabel={RMSE},
ytick ={0,0.2,0.4,0.6,0.8,1.0,1.2,1.4,1.6,1.8},
axis background/.style={fill=white},
legend style={legend cell align=left, align=left,draw=white!15!black},
legend pos=north east
]

\addplot [color=mycolor4,line width=0.75pt]
  table[row sep=crcr]{%
0	0	\\
0.2	0.86	\\
0.4	0.90	\\
0.6	0.98	\\
0.8	1.06	\\
1.0	1.10	\\
1.2	1.16	\\
1.4	1.21	\\
};
\addlegendentry{FGSM}

\addplot [color=mycolor1,line width=0.75pt]
  table[row sep=crcr]{%
0	0	\\
0.2	0.98	\\
0.4	1.08	\\
0.6	1.25	\\
0.8	1.32	\\
1.0	1.45	\\
1.2	1.58	\\
1.4	1.62	\\
};
\addlegendentry{BIM}

\end{axis}
\end{tikzpicture}%
}\label{fig:GoogleEps}}
	\end{subfigure}~
	\caption{RMSE variation with respect to the amount of perturbation ($\epsilon$) for FGSM and BIM attacks }\label{fig:Epsilonrmse}
\end{figure}
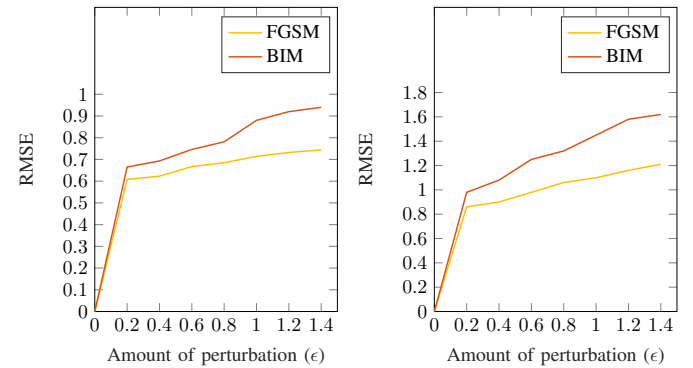

In \figurename~\ref{fig:Epsilonrmse}, we evaluate the LSTM and GRU regression model's performance with respect to the different amount of perturbations allowed for crafting the adversarial MTS examples. We pick the LSTM and GRU as they showed the best performance for the MTS regression task in \figurename~\ref{fig:Power_comp} and \figurename~\ref{fig:Google_comp}. We observe that for larger values of $\epsilon$, FGSM is not very helpful in generating adversarial MTS examples for fooling the LSTM and GRU regression model. In comparison, with larger values of $\epsilon$, BIM crafts more devastating adversarial MTS examples for fooling both the regression models and thus RMSE follows an increasing trend. This is due to the fact~\cite{kurakin2016adversarial} that BIM adds a small amount of perturbation $\alpha$ on each iteration whereas FGSM adds $\epsilon$ amount of noise for each data point in the MTS that may not be very helpful in generating inaccurate forecasting with higher RMSE values.  

\begin{table*}[ht!]
\centering
\caption {Transferability of FGSM and BIM attacks for power Consumption and Google stock datasets. The notation X/Y represents the percentage of RMSE increase using FGSM/BIM}

\begin{tabular}{|c|c|c|c||c|c|c|}
\hline
\multirow{3}{*}{\begin{tabular}[c]{@{}c@{}}\textbf{DL} \\ \textbf{models}\end{tabular}} & \multicolumn{6}{c|}{\textbf{Transferability (\% increase of RMSE)}}                                                                                                                                      \\ \cline{2-7} 
                                                                      & \multicolumn{3}{c||}{\textbf{Power consumption dataset}}                                                 & \multicolumn{3}{c|}{\textbf{Google stock dataset} }                                                     \\ \cline{2-7}  & \textbf{CNN}   & \textbf{LSTM} & \textbf{GRU}  & \textbf{CNN}   & \textbf{LSTM} & \textbf{GRU}                                       \\  \hline
CNN   & - & 10.2/18.7 & 10.8/18.1  & - & 16.9/24.1 & 16.2/23.4  \\ [0.5ex] 
LSTM  & 8.3/16.9 & - & 7.5/11.2  & 13.1/18.6 & - & 11.1/16.4  \\ [0.5ex]
GRU   & 9.2/16.5 & 6.6/11.7 & -  & 13.8/19.7 & 11.6/16.3 & -  \\ [0.5ex]  
 \hline
\end{tabular}
\label{tab:Tran_table}

\end{table*}

\subsection{Transferability of adversarial examples}
To evaluate the transferability of adversarial attacks, we apply the adversarial examples crafted for a DL MTS regression model on the other DL models. Table~\ref{tab:Tran_table} summarizes the obtained results on transferability. We observe that for both datasets, the adversarial examples crafted for CNN are the most transferable. This means a higher RMSE is observed when adversarial examples crafted for the CNN model are transferred to other models. For instance, adversarial MTS examples crafted using BIM for the CNN regression model (Google stock dataset) causes a 23.4\% increase when transferred to the GRU regression model. A similar trend is also observed, however, with a lower percentage increases, when adversarial examples crafted for GRU and LSTM regression models are transferred to the other DL regression models. In addition, the obtained results also show that BIM is better than FGSM in fooling (even when they are transferred) the DL models for MTS regression tasks, e.g. BIM increases the RMSE more when compared to the FGSM. Overall, the results show that the adversarial examples are capable of generalizing to a different DL network architecture. This type of attack is known as \emph{black box} attacks, where the attackers do not have access to the target model’s internal parameters, yet they are able to generate perturbed time series that fool the DL models for MTSR tasks.

\subsection{Defense against adversarial attacks}
Researchers have proposed different types of adversarial attack defense strategies so far~\cite{qiu2019review} most of which are applicable to the image domain. The existing adversarial attack defense strategies can be divided into three categories: \emph{modifying data}, \emph{modifying models}, and \emph{using auxiliary tools}. Modifying data refers to modifying the training dataset in the training stage, or changing the input data in the testing stage. It also includes adversarial training~\cite{goodfellow2014explaining}, blocking the transferability~\cite{hosseini2017blocking}, data compression~\cite{das2017keeping}, gradient hiding~\cite{papernot2017practical}, and data randomization~\cite{xie2017adversarial}. In contrast, modifying models refer to the modification of DL models, such as defensive distillation~\cite{papernot2016distillation}, feature squeezing~\cite{Tablefeature}, regularization~\cite{biggio2011support}, deep contractive network~\cite{gu2014towards} and mask defense~\cite{gao2017deepcloak}. Using additional tools to the DL models is referred to as using an auxiliary tool which includes the use of defense-GAN~\cite{TableGan}, MagNet~\cite{meng2017magnet} and high-level representation guided denoiser~\cite{liao2018defense}. Unfortunately, most of these detectors are prone to adversarial attacks due to the fact that these attacks are designed specifically to fool such detectors~\cite{yuan2019adversarial}. Hence, the time series, data mining and machine learning need to pay special attention to this area as DL MTS regression models are gaining popularity in the safety and cost-critical application domains. A potential idea for the detection of adversarial examples in MTS DL regression models can be the use of inductive conformal anomaly detection method~\cite{balasubramanian2014conformal,volkhonskiy2017inductive}. Another potential idea is to leverage the decades of research into non-probabilistic classifiers, such as the nearest neighbor coupled with DTW~\cite{yuan2019adversarial}.

\section{Conclusion}


In this paper, we introduced the concept of adversarial attacks on deep learning (DL) regression models for multivariate time series (MTS) regression. We formalized and evaluated two adversarial example generation techniques, originally proposed for the image domain for the MTS regression task. The obtained results showed how adversarial attacks can induce inaccurate forecasting when evaluated on the household power consumption and the Google stock dataset. We also observed that BIM is not only a more stealthy attack but also causes higher damage in DL MTS regression models. Finally, among the three evaluated DL regression models, the obtained results revealed that the adversarial examples crafted for CNN are more transferable when compared to the others. Through our work, we shed light on the importance of acknowledging adversarial attacks as one of the prominent threats to the DL MTS regression models as they find their applications in safety-critical and cost-critical domains. 

In the future, we would like to extend our work by adapting other adversarial attacks for the image domain and evaluate them for MTS DL regression. In addition, we also plan to investigate defense strategies to detect and mitigate adversarial threats in DL regression models.

\bibliographystyle{IEEEtran}
\bibliography{ref}

\begin{thebibliography}{10}
\providecommand{\url}[1]{#1}
\csname url@samestyle\endcsname
\providecommand{\newblock}{\relax}
\providecommand{\bibinfo}[2]{#2}
\providecommand{\BIBentrySTDinterwordspacing}{\spaceskip=0pt\relax}
\providecommand{\BIBentryALTinterwordstretchfactor}{4}
\providecommand{\BIBentryALTinterwordspacing}{\spaceskip=\fontdimen2\font plus
\BIBentryALTinterwordstretchfactor\fontdimen3\font minus
  \fontdimen4\font\relax}
\providecommand{\BIBforeignlanguage}[2]{{%
\expandafter\ifx\csname l@#1\endcsname\relax
\typeout{** WARNING: IEEEtran.bst: No hyphenation pattern has been}%
\typeout{** loaded for the language `#1'. Using the pattern for}%
\typeout{** the default language instead.}%
\else
\language=\csname l@#1\endcsname
\fi
#2}}
\providecommand{\BIBdecl}{\relax}
\BIBdecl

\bibitem{sezer2020financial}
O.~B. Sezer, M.~U. Gudelek, and A.~M. Ozbayoglu, ``Financial time series
  forecasting with deep learning: A systematic literature review: 2005--2019,''
  \emph{Applied Soft Computing}, vol.~90, p. 106181, 2020.

\bibitem{gan2020machine}
L.~Gan, H.~Wang, and Z.~Yang, ``Machine learning solutions to challenges in
  finance: An application to the pricing of financial products,''
  \emph{Technological Forecasting and Social Change}, vol. 153, p. 119928,
  2020.

\bibitem{sirignano2019universal}
J.~Sirignano and R.~Cont, ``Universal features of price formation in financial
  markets: perspectives from deep learning,'' \emph{Quantitative Finance},
  vol.~19, no.~9, pp. 1449--1459, 2019.

\bibitem{lee2019multimodal}
S.~I. Lee and S.~J. Yoo, ``Multimodal deep learning for finance: integrating
  and forecasting international stock markets,'' \emph{The Journal of
  Supercomputing}, pp. 1--19, 2019.

\bibitem{salman2015weather}
A.~G. Salman, B.~Kanigoro, and Y.~Heryadi, ``Weather forecasting using deep
  learning techniques,'' in \emph{2015 international conference on advanced
  computer science and information systems (ICACSIS)}.\hskip 1em plus 0.5em
  minus 0.4em\relax IEEE, 2015, pp. 281--285.

\bibitem{hossain2015forecasting}
M.~Hossain, B.~Rekabdar, S.~J. Louis, and S.~Dascalu, ``Forecasting the weather
  of nevada: A deep learning approach,'' in \emph{2015 international joint
  conference on neural networks (IJCNN)}.\hskip 1em plus 0.5em minus
  0.4em\relax IEEE, 2015, pp. 1--6.

\bibitem{khan2018load}
M.~Khan, N.~Javaid, M.~N. Iqbal, M.~Bilal, S.~F.~A. Zaidi, and R.~A. Raza,
  ``Load prediction based on multivariate time series forecasting for energy
  consumption and behavioral analytics,'' in \emph{Conference on Complex,
  Intelligent, and Software Intensive Systems}.\hskip 1em plus 0.5em minus
  0.4em\relax Springer, 2018, pp. 305--316.

\bibitem{chan2019deep}
S.~Chan, I.~Oktavianti, and V.~Puspita, ``A deep learning cnn and ai-tuned svm
  for electricity consumption forecasting: Multivariate time series data,'' in
  \emph{2019 IEEE 10th Annual Information Technology, Electronics and Mobile
  Communication Conference (IEMCON)}.\hskip 1em plus 0.5em minus 0.4em\relax
  IEEE, 2019, pp. 0488--0494.

\bibitem{fontes2016pattern}
C.~H. Fontes and O.~Pereira, ``Pattern recognition in multivariate time
  series--a case study applied to fault detection in a gas turbine,''
  \emph{Engineering Applications of Artificial Intelligence}, vol.~49, pp.
  10--18, 2016.

\bibitem{lei2019fault}
J.~Lei, C.~Liu, and D.~Jiang, ``Fault diagnosis of wind turbine based on long
  short-term memory networks,'' \emph{Renewable energy}, vol. 133, pp.
  422--432, 2019.

\bibitem{zou2018towards}
H.~Zou, Y.~Zhou, J.~Yang, and C.~J. Spanos, ``Towards occupant activity driven
  smart buildings via wifi-enabled iot devices and deep learning,''
  \emph{Energy and Buildings}, vol. 177, pp. 12--22, 2018.

\bibitem{zhang2018thermal}
W.~Zhang, W.~Hu, and Y.~Wen, ``Thermal comfort modeling for smart buildings: A
  fine-grained deep learning approach,'' \emph{IEEE Internet of Things
  Journal}, vol.~6, no.~2, pp. 2540--2549, 2018.

\bibitem{fawaz2019deep}
H.~I. Fawaz, G.~Forestier, J.~Weber, L.~Idoumghar, and P.-A. Muller, ``Deep
  neural network ensembles for time series classification,'' in \emph{2019
  International Joint Conference on Neural Networks (IJCNN)}.\hskip 1em plus
  0.5em minus 0.4em\relax IEEE, 2019, pp. 1--6.

\bibitem{goodfellow2014explaining}
I.~J. Goodfellow, J.~Shlens, and C.~Szegedy, ``Explaining and harnessing
  adversarial examples,'' \emph{arXiv preprint arXiv:1412.6572}, 2014.

\bibitem{kurakin2016adversarial}
A.~Kurakin, I.~Goodfellow, and S.~Bengio, ``Adversarial examples in the
  physical world,'' \emph{arXiv preprint arXiv:1607.02533}, 2016.

\bibitem{adrel2}
C.~Szegedy, W.~Zaremba, I.~Sutskever, J.~Bruna, D.~Erhan, I.~Goodfellow, and
  R.~Fergus, ``Intriguing properties of neural networks,'' \emph{arXiv preprint
  arXiv:1312.6199}, 2013.

\bibitem{adrel7}
Y.~Liu, X.~Chen, C.~Liu, and D.~Song, ``Delving into transferable adversarial
  examples and black-box attacks,'' \emph{arXiv preprint arXiv:1611.02770},
  2016.

\bibitem{mitrel1}
A.~Goel, A.~Singh, A.~Agarwal, M.~Vatsa, and R.~Singh, ``Smartbox: Benchmarking
  adversarial detection and mitigation algorithms for face recognition,'' in
  \emph{2018 IEEE 9th International Conference on Biometrics Theory,
  Applications and Systems (BTAS)}.\hskip 1em plus 0.5em minus 0.4em\relax
  IEEE, 2018, pp. 1--7.

\bibitem{mitrel2}
I.~Rosenberg, A.~Shabtai, Y.~Elovici, and L.~Rokach, ``Defense methods against
  adversarial examples for recurrent neural networks,'' \emph{arXiv preprint
  arXiv:1901.09963}, 2019.

\bibitem{mitrel3}
G.~Goswami, N.~Ratha, A.~Agarwal, R.~Singh, and M.~Vatsa, ``Unravelling
  robustness of deep learning based face recognition against adversarial
  attacks,'' in \emph{Thirty-Second AAAI Conference on Artificial
  Intelligence}, 2018.

\bibitem{mitrel4}
S.~Kokalj-Filipovic, R.~Miller, N.~Chang, and C.~L. Lau, ``Mitigation of
  adversarial examples in rf deep classifiers utilizing autoencoder
  pre-training,'' \emph{arXiv preprint arXiv:1902.08034}, 2019.

\bibitem{mitrel5}
L.~Song, R.~Shokri, and P.~Mittal, ``Privacy risks of securing machine learning
  models against adversarial examples,'' \emph{arXiv preprint
  arXiv:1905.10291}, 2019.

\bibitem{mitrel6}
C.~Song, H.-P. Cheng, H.~Yang, S.~Li, C.~Wu, Q.~Wu, Y.~Chen, and H.~Li, ``Mat:
  A multi-strength adversarial training method to mitigate adversarial
  attacks,'' in \emph{2018 IEEE Computer Society Annual Symposium on VLSI
  (ISVLSI)}.\hskip 1em plus 0.5em minus 0.4em\relax IEEE, 2018, pp. 476--481.

\bibitem{carlini2017provably}
N.~Carlini, G.~Katz, C.~Barrett, and D.~L. Dill, ``Provably minimally-distorted
  adversarial examples,'' \emph{arXiv preprint arXiv:1709.10207}, 2017.

\bibitem{rival2019}
H.~I. Fawaz, G.~Forestier, J.~Weber, L.~Idoumghar, and P.-A. Muller,
  ``Adversarial attacks on deep neural networks for time series
  classification,'' in \emph{2019 International Joint Conference on Neural
  Networks (IJCNN)}.\hskip 1em plus 0.5em minus 0.4em\relax IEEE, 2019, pp.
  1--8.

\bibitem{wang2017time}
Z.~Wang, W.~Yan, and T.~Oates, ``Time series classification from scratch with
  deep neural networks: A strong baseline,'' in \emph{2017 International joint
  conference on neural networks (IJCNN)}.\hskip 1em plus 0.5em minus
  0.4em\relax IEEE, 2017, pp. 1578--1585.

\bibitem{musleh2019survey}
A.~S. Musleh, G.~Chen, and Z.~Y. Dong, ``A survey on the detection algorithms
  for false data injection attacks in smart grids,'' \emph{IEEE Transactions on
  Smart Grid}, 2019.

\bibitem{gasparin2019deep}
A.~Gasparin, S.~Lukovic, and C.~Alippi, ``Deep learning for time series
  forecasting: The electric load case,'' \emph{arXiv preprint
  arXiv:1907.09207}, 2019.

\bibitem{LSTM-org11}
S.~Hochreiter and J.~Schmidhuber, ``Lstm can solve hard long time lag
  problems,'' in \emph{Advances in neural information processing systems},
  1997, pp. 473--479.

\bibitem{GRU-org}
K.~Cho, B.~Van~Merri{\"e}nboer, C.~Gulcehre, D.~Bahdanau, F.~Bougares,
  H.~Schwenk, and Y.~Bengio, ``Learning phrase representations using rnn
  encoder-decoder for statistical machine translation,'' \emph{arXiv preprint
  arXiv:1406.1078}, 2014.

\bibitem{CNN-org}
J.~Gu, Z.~Wang, J.~Kuen, L.~Ma, A.~Shahroudy, B.~Shuai, T.~Liu, X.~Wang,
  G.~Wang, J.~Cai \emph{et~al.}, ``Recent advances in convolutional neural
  networks,'' \emph{Pattern Recognition}, vol.~77, pp. 354--377, 2018.

\bibitem{GoogleDataset}
\BIBentryALTinterwordspacing
{www.nasdaq.com}, ``Google stock dataset (www.nasdaq.com),'' 2020, [Online;
  accessed 04-March-2020]. [Online]. Available:
  \url{https://www.nasdaq.com/symbol/goog/historical}
\BIBentrySTDinterwordspacing

\bibitem{PowerDataset}
\BIBentryALTinterwordspacing
G.~Hebrail, ``Individual household electric power consumption data set,'' 2012,
  [Online; accessed 15-March-2020]. [Online]. Available:
  \url{https://archive.ics.uci.edu/ml/datasets/individual+household+electric+power+consumption}
\BIBentrySTDinterwordspacing

\bibitem{de200625}
J.~G. De~Gooijer and R.~J. Hyndman, ``25 years of time series forecasting,''
  \emph{International journal of forecasting}, vol.~22, no.~3, pp. 443--473,
  2006.

\bibitem{naing2015state}
N.~Naing, W.~Yan, and Z.~Z. Htike, ``State of the art machine learning
  techniques for time series forecasting: A survey,'' \emph{Advanced Science
  Letters}, vol.~21, no.~11, pp. 3574--3576, 2015.

\bibitem{tealab2018time}
A.~Tealab, ``Time series forecasting using artificial neural networks
  methodologies: A systematic review,'' \emph{Future Computing and Informatics
  Journal}, vol.~3, no.~2, pp. 334--340, 2018.

\bibitem{borovykh2017conditional}
A.~Borovykh, S.~Bohte, and C.~W. Oosterlee, ``Conditional time series
  forecasting with convolutional neural networks,'' \emph{arXiv preprint
  arXiv:1703.04691}, 2017.

\bibitem{gamboa2017deep}
J.~C.~B. Gamboa, ``Deep learning for time-series analysis,'' \emph{arXiv
  preprint arXiv:1701.01887}, 2017.

\bibitem{vengertsev2014deep}
D.~Vengertsev, ``Deep learning architecture for univariate time series
  forecasting,'' \emph{Cs229}, pp. 3--7, 2014.

\bibitem{tian2018lstm}
Y.~Tian, K.~Zhang, J.~Li, X.~Lin, and B.~Yang, ``Lstm-based traffic flow
  prediction with missing data,'' \emph{Neurocomputing}, vol. 318, pp.
  297--305, 2018.

\bibitem{sagheer2019time}
A.~Sagheer and M.~Kotb, ``Time series forecasting of petroleum production using
  deep lstm recurrent networks,'' \emph{Neurocomputing}, vol. 323, pp.
  203--213, 2019.

\bibitem{cao2019financial}
J.~Cao, Z.~Li, and J.~Li, ``Financial time series forecasting model based on
  ceemdan and lstm,'' \emph{Physica A: Statistical Mechanics and its
  Applications}, vol. 519, pp. 127--139, 2019.

\bibitem{sorkun2020time}
M.~C. Sorkun, {\"O}.~D. {\.I}NCEL, and C.~Paoli, ``Time series forecasting on
  multivariate solar radiation data using deep learning (lstm),'' \emph{Turkish
  Journal of Electrical Engineering \& Computer Sciences}, vol.~28, no.~1, pp.
  211--223, 2020.

\bibitem{yuan2016fault}
M.~Yuan, Y.~Wu, and L.~Lin, ``Fault diagnosis and remaining useful life
  estimation of aero engine using lstm neural network,'' in \emph{2016 IEEE
  International Conference on Aircraft Utility Systems (AUS)}.\hskip 1em plus
  0.5em minus 0.4em\relax IEEE, 2016, pp. 135--140.

\bibitem{RNN-org}
R.~Jozefowicz, W.~Zaremba, and I.~Sutskever, ``An empirical exploration of
  recurrent network architectures,'' in \emph{International conference on
  machine learning}, 2015, pp. 2342--2350.

\bibitem{yamak2019comparison}
P.~T. Yamak, L.~Yujian, and P.~K. Gadosey, ``A comparison between arima, lstm,
  and gru for time series forecasting,'' in \emph{Proceedings of the 2019 2nd
  International Conference on Algorithms, Computing and Artificial
  Intelligence}, 2019, pp. 49--55.

\bibitem{tao2019air}
Q.~Tao, F.~Liu, Y.~Li, and D.~Sidorov, ``Air pollution forecasting using a deep
  learning model based on 1d convnets and bidirectional gru,'' \emph{IEEE
  Access}, vol.~7, pp. 76\,690--76\,698, 2019.

\bibitem{de2019gru}
E.~De~Brouwer, J.~Simm, A.~Arany, and Y.~Moreau, ``Gru-ode-bayes: Continuous
  modeling of sporadically-observed time series,'' in \emph{Advances in Neural
  Information Processing Systems}, 2019, pp. 7377--7388.

\bibitem{jia2020research}
P.~Jia, H.~Liu, S.~Wang, and P.~Wang, ``Research on a mine gas concentration
  forecasting model based on a gru network,'' \emph{IEEE Access}, vol.~8, pp.
  38\,023--38\,031, 2020.

\bibitem{shen2020short}
M.~Shen, Q.~Xu, K.~Wang, M.~Tu, and B.~Wu, ``Short-term bus load forecasting
  method based on cnn-gru neural network,'' in \emph{Proceedings of PURPLE
  MOUNTAIN FORUM 2019-International Forum on Smart Grid Protection and
  Control}.\hskip 1em plus 0.5em minus 0.4em\relax Springer, 2020, pp.
  711--722.

\bibitem{dong2017cnn}
X.~Dong, L.~Qian, and L.~Huang, ``A cnn based bagging learning approach to
  short-term load forecasting in smart grid,'' in \emph{2017 IEEE SmartWorld,
  Ubiquitous Intelligence \& Computing, Advanced \& Trusted Computed, Scalable
  Computing \& Communications, Cloud \& Big Data Computing, Internet of People
  and Smart City Innovation (SmartWorld/SCALCOM/UIC/ATC/CBDCom/IOP/SCI)}.\hskip
  1em plus 0.5em minus 0.4em\relax IEEE, 2017, pp. 1--6.

\bibitem{arratia2019convolutional}
A.~Arratia and E.~Sep{\'u}lveda, ``Convolutional neural networks, image
  recognition and financial time series forecasting,'' in \emph{Workshop on
  Mining Data for Financial Applications}.\hskip 1em plus 0.5em minus
  0.4em\relax Springer, 2019, pp. 60--69.

\bibitem{zheng2017wide}
Z.~Zheng, Y.~Yang, X.~Niu, H.-N. Dai, and Y.~Zhou, ``Wide and deep
  convolutional neural networks for electricity-theft detection to secure smart
  grids,'' \emph{IEEE Transactions on Industrial Informatics}, vol.~14, no.~4,
  pp. 1606--1615, 2017.

\bibitem{mode2020impact}
G.~R. Mode, P.~Calyam, and K.~A. Hoque, ``Impact of false data injection
  attacks on deep learning enabled predictive analytics,'' in \emph{NOMS
  2020-2020 IEEE/IFIP Network Operations and Management Symposium}.\hskip 1em
  plus 0.5em minus 0.4em\relax IEEE, 2020, pp. 1--7.

\bibitem{ngai2011application}
E.~W. Ngai, Y.~Hu, Y.~H. Wong, Y.~Chen, and X.~Sun, ``The application of data
  mining techniques in financial fraud detection: A classification framework
  and an academic review of literature,'' \emph{Decision support systems},
  vol.~50, no.~3, pp. 559--569, 2011.

\bibitem{das2012stock}
S.~Das, A.~Mukhopadhyay, and M.~Anand, ``Stock market response to information
  security breach: A study using firm and attack characteristics,''
  \emph{Journal of Information Privacy and Security}, vol.~8, no.~4, pp.
  27--55, 2012.

\bibitem{akshaya2019taxonomy}
M.~S. Akshaya and G.~Padmavathi, ``Taxonomy of security attacks and risk
  assessment of cloud computing,'' in \emph{Advances in Big Data and Cloud
  Computing}.\hskip 1em plus 0.5em minus 0.4em\relax Springer, 2019, pp.
  37--59.

\bibitem{szegedy2013intriguing}
C.~Szegedy, W.~Zaremba, I.~Sutskever, J.~Bruna, D.~Erhan, I.~Goodfellow, and
  R.~Fergus, ``Intriguing properties of neural networks,'' \emph{arXiv preprint
  arXiv:1312.6199}, 2013.

\bibitem{TeslaHack}
\BIBentryALTinterwordspacing
I.~A. Hamilton, ``A 2-inch strip of tape on a 35-mph speed sign and
  successfully tricked 2 teslas into accelerating to 85 mph,'' 2020, [Online;
  accessed 06-March-2020]. [Online]. Available:
  \url{https://www.businessinsider.com/hackers-trick-tesla-accelerating-85mph-using-tape-2020-2}
\BIBentrySTDinterwordspacing

\bibitem{adrel5}
A.~Madry, A.~Makelov, L.~Schmidt, D.~Tsipras, and A.~Vladu, ``Towards deep
  learning models resistant to adversarial attacks,'' \emph{arXiv preprint
  arXiv:1706.06083}, 2017.

\bibitem{qiu2019review}
S.~Qiu, Q.~Liu, S.~Zhou, and C.~Wu, ``Review of artificial intelligence
  adversarial attack and defense technologies,'' \emph{Applied Sciences},
  vol.~9, no.~5, p. 909, 2019.

\bibitem{xu2019adversarial}
H.~Xu, Y.~Ma, H.~Liu, D.~Deb, H.~Liu, J.~Tang, and A.~Jain, ``Adversarial
  attacks and defenses in images, graphs and text: A review,'' \emph{arXiv
  preprint arXiv:1909.08072}, 2019.

\bibitem{biggio2015adversarial}
B.~Biggio, P.~Russu, L.~Didaci, F.~Roli \emph{et~al.}, ``Adversarial biometric
  recognition: A review on biometric system security from the adversarial
  machine-learning perspective,'' \emph{IEEE Signal Processing Magazine},
  vol.~32, no.~5, pp. 31--41, 2015.

\bibitem{yuan2019adversarial}
X.~Yuan, P.~He, Q.~Zhu, and X.~Li, ``Adversarial examples: Attacks and defenses
  for deep learning,'' \emph{IEEE transactions on neural networks and learning
  systems}, vol.~30, no.~9, pp. 2805--2824, 2019.

\bibitem{oregi2018adversarial}
I.~Oregi, J.~Del~Ser, A.~Perez, and J.~A. Lozano, ``Adversarial sample crafting
  for time series classification with elastic similarity measures,'' in
  \emph{International Symposium on Intelligent and Distributed
  Computing}.\hskip 1em plus 0.5em minus 0.4em\relax Springer, 2018, pp.
  26--39.

\bibitem{bagnall2017great}
A.~Bagnall, J.~Lines, A.~Bostrom, J.~Large, and E.~Keogh, ``The great time
  series classification bake off: a review and experimental evaluation of
  recent algorithmic advances,'' \emph{Data Mining and Knowledge Discovery},
  vol.~31, no.~3, pp. 606--660, 2017.

\bibitem{kim2019predicting}
T.-Y. Kim and S.-B. Cho, ``Predicting residential energy consumption using
  cnn-lstm neural networks,'' \emph{Energy}, vol. 182, pp. 72--81, 2019.

\bibitem{wang2019data}
Z.~Wang, T.~Hong, and M.~A. Piette, ``Data fusion in predicting internal heat
  gains for office buildings through a deep learning approach,'' \emph{Applied
  energy}, vol. 240, pp. 386--398, 2019.

\bibitem{moon2018forecasting}
J.~Moon, J.~Park, E.~Hwang, and S.~Jun, ``Forecasting power consumption for
  higher educational institutions based on machine learning,'' \emph{The
  Journal of Supercomputing}, vol.~74, no.~8, pp. 3778--3800, 2018.

\bibitem{chai2014root}
T.~Chai and R.~R. Draxler, ``Root mean square error (rmse) or mean absolute
  error (mae)?--arguments against avoiding rmse in the literature,''
  \emph{Geoscientific model development}, vol.~7, no.~3, pp. 1247--1250, 2014.

\bibitem{kingma2014adam}
D.~P. Kingma and J.~Ba, ``Adam: A method for stochastic optimization,''
  \emph{arXiv preprint arXiv:1412.6980}, 2014.

\bibitem{zoller2019survey}
M.-A. Z{\"o}ller and M.~F. Huber, ``Survey on automated machine learning,''
  \emph{arXiv preprint arXiv:1904.12054}, 2019.

\bibitem{long2019deep}
W.~Long, Z.~Lu, and L.~Cui, ``Deep learning-based feature engineering for stock
  price movement prediction,'' \emph{Knowledge-Based Systems}, vol. 164, pp.
  163--173, 2019.

\bibitem{song2019study}
Y.~Song, J.~W. Lee, and J.~Lee, ``A study on novel filtering and relationship
  between input-features and target-vectors in a deep learning model for stock
  price prediction,'' \emph{Applied Intelligence}, vol.~49, no.~3, pp.
  897--911, 2019.

\bibitem{hosseini2017blocking}
H.~Hosseini, Y.~Chen, S.~Kannan, B.~Zhang, and R.~Poovendran, ``Blocking
  transferability of adversarial examples in black-box learning systems,''
  \emph{arXiv preprint arXiv:1703.04318}, 2017.

\bibitem{das2017keeping}
N.~Das, M.~Shanbhogue, S.-T. Chen, F.~Hohman, L.~Chen, M.~E. Kounavis, and
  D.~H. Chau, ``Keeping the bad guys out: Protecting and vaccinating deep
  learning with jpeg compression,'' \emph{arXiv preprint arXiv:1705.02900},
  2017.

\bibitem{papernot2017practical}
N.~Papernot, P.~McDaniel, I.~Goodfellow, S.~Jha, Z.~B. Celik, and A.~Swami,
  ``Practical black-box attacks against machine learning,'' in
  \emph{Proceedings of the 2017 ACM on Asia conference on computer and
  communications security}.\hskip 1em plus 0.5em minus 0.4em\relax ACM, 2017,
  pp. 506--519.

\bibitem{xie2017adversarial}
C.~Xie, J.~Wang, Z.~Zhang, Y.~Zhou, L.~Xie, and A.~Yuille, ``Adversarial
  examples for semantic segmentation and object detection,'' in
  \emph{Proceedings of the IEEE International Conference on Computer Vision},
  2017, pp. 1369--1378.

\bibitem{papernot2016distillation}
N.~Papernot, P.~McDaniel, X.~Wu, S.~Jha, and A.~Swami, ``Distillation as a
  defense to adversarial perturbations against deep neural networks,'' in
  \emph{2016 IEEE Symposium on Security and Privacy (SP)}.\hskip 1em plus 0.5em
  minus 0.4em\relax IEEE, 2016, pp. 582--597.

\bibitem{Tablefeature}
W.~Xu, D.~Evans, and Y.~Qi, ``Feature squeezing: Detecting adversarial examples
  in deep neural networks,'' \emph{arXiv preprint arXiv:1704.01155}, 2017.

\bibitem{biggio2011support}
B.~Biggio, B.~Nelson, and P.~Laskov, ``Support vector machines under
  adversarial label noise,'' in \emph{Asian conference on machine learning},
  2011, pp. 97--112.

\bibitem{gu2014towards}
S.~Gu and L.~Rigazio, ``Towards deep neural network architectures robust to
  adversarial examples,'' \emph{arXiv preprint arXiv:1412.5068}, 2014.

\bibitem{gao2017deepcloak}
J.~Gao, B.~Wang, Z.~Lin, W.~Xu, and Y.~Qi, ``Deepcloak: Masking deep neural
  network models for robustness against adversarial samples,'' \emph{arXiv
  preprint arXiv:1702.06763}, 2017.

\bibitem{TableGan}
P.~Samangouei, M.~Kabkab, and R.~Chellappa, ``Defense-gan: Protecting
  classifiers against adversarial attacks using generative models,''
  \emph{arXiv preprint arXiv:1805.06605}, 2018.

\bibitem{meng2017magnet}
D.~Meng and H.~Chen, ``Magnet: a two-pronged defense against adversarial
  examples,'' in \emph{Proceedings of the 2017 ACM SIGSAC Conference on
  Computer and Communications Security}.\hskip 1em plus 0.5em minus 0.4em\relax
  ACM, 2017, pp. 135--147.

\bibitem{liao2018defense}
F.~Liao, M.~Liang, Y.~Dong, T.~Pang, X.~Hu, and J.~Zhu, ``Defense against
  adversarial attacks using high-level representation guided denoiser,'' in
  \emph{Proceedings of the IEEE Conference on Computer Vision and Pattern
  Recognition}, 2018, pp. 1778--1787.

\bibitem{balasubramanian2014conformal}
V.~Balasubramanian, S.-S. Ho, and V.~Vovk, \emph{Conformal prediction for
  reliable machine learning: theory, adaptations and applications}.\hskip 1em
  plus 0.5em minus 0.4em\relax Newnes, 2014.

\bibitem{volkhonskiy2017inductive}
D.~Volkhonskiy, I.~Nouretdinov, A.~Gammerman, V.~Vovk, and E.~Burnaev,
  ``Inductive conformal martingales for change-point detection,'' \emph{arXiv
  preprint arXiv:1706.03415}, 2017.

\end{thebibliography}

\end{document}